\documentclass{article}

\usepackage{neurips_2025}

\usepackage{graphicx}
\usepackage{multirow}
\usepackage{float}

\usepackage[utf8]{inputenc} 
\usepackage[T1]{fontenc}    
\usepackage{hyperref}       
\usepackage{url}            
\usepackage{booktabs}       
\usepackage{amsfonts}       
\usepackage{nicefrac}       
\usepackage{microtype}      
\usepackage{enumitem}
\usepackage{caption}
\usepackage[list=true]{subcaption}       
\usepackage[table]{xcolor}    
\usepackage{colortbl}         
\usepackage{multirow}         
\usepackage{rotating}         
\usepackage{geometry}         
\usepackage{graphicx}
\usepackage{adjustbox}
\usepackage{amsmath}
\usepackage{arydshln}
\usepackage{threeparttable}
\usepackage[most]{tcolorbox} 
\tcbuselibrary{listingsutf8} 
\usepackage{siunitx} 
\usepackage{array}            
\usepackage[at]{easylist}
\usepackage{todonotes}
\nolinenumbers
\definecolor{headercolor}{gray}{0.55}
\definecolor{headergray}{gray}{0.75}
\definecolor{sectiongray}{gray}{0.90}
\definecolor{rowgray}{gray}{0.90} 
\definecolor{lightgray}{gray}{0.95} 
\definecolor{tablegray}{gray}{0.9} 

\definecolor{lightred}{RGB}{230, 85, 85}
\definecolor{lightblue}{RGB}{80, 150, 230} 

\newcommand{\stackth}[1]{\textbf{\shortstack{#1}}}

\definecolor{headergreen}{rgb}{0.9, 1.0, 0.9}      
\definecolor{emuyellow}{rgb}{1.0, 0.95, 0.7}      
\definecolor{magicbrushorange}{rgb}{1.0, 0.85, 0.7} 
\definecolor{redtext}{rgb}{1.0, 0.0, 0.0}         

\definecolor{headerblue}{HTML}{004F6E}
\definecolor{lightred}{HTML}{D9534F}
\definecolor{lightgreen}{HTML}{5CB85C}

\newcommand{\redx}{\textcolor{lightred}{$\times$}}
\newcommand{\greencheck}{\textcolor{lightgreen}{$\checkmark$}}

\renewrobustcmd{\bfseries}{\fontseries{b}\selectfont}
\renewrobustcmd{\boldmath}{}
\title{MADI: Masking-Augmented Diffusion with Inference-Time Scaling for Visual Editing}

\author{%
  Shreya Kadambi \thanks{These authors contributed equally to this work.}  \quad Risheek Garrepalli$^*$ \quad Shubhankar Borse \quad Munawar Hyatt \quad
  \textbf{Fatih Porikli}\\ 
  Qualcomm AI Research\thanks{Qualcomm AI Research is an initiative of Qualcomm Technologies, Inc.}\\
  \texttt{\{skadambi, rgarrepa, sborse, mhayat, fporikli\}@qti.qualcomm.com}\\ 
}
\begin{document}

\maketitle

\begin{abstract}

Despite the remarkable success of diffusion models in text-to-image generation, their effectiveness in grounded visual editing and compositional control remains challenging. Motivated by advances in self-supervised learning and in-context generative modeling, we propose a series of simple yet powerful design choices that significantly enhance diffusion model’s capacity for structured, controllable generation and editing. We introduce \textit{Masking-Augmented Diffusion with Inference-Time Scaling (MADI)}, a framework that improves the editability, compositionality and controllability of diffusion models through two core innovations. 
First, we introduce \textbf{Masking-Augmented gaussian Diffusion (MAgD)}, a novel training strategy  with dual corruption process which combines standard denoising score matching and masked reconstruction by masking noisy input from forward process. 
MAgD encourages the model to learn discriminative and compositional visual representations, thus enabling localized and structure-aware editing. Second, we introduce an \textbf{inference-time capacity scaling} mechanism based on \textbf{Pause Tokens}, which act as special placeholders inserted into the prompt for increasing computational capacity at inference time. 
Our findings show that adopting expressive and dense prompts during training further enhances performance, particularly for MAgD.
Together, these contributions in MADI substantially enhance the \textit{editability} of diffusion models, paving the way toward their integration into more \textit{general-purpose, in-context generative diffusion architectures}.


\end{abstract}

\section{Introduction}

Diffusion-based generative models \cite{ho2020denoising, song2020scorebased,rombach2022high} have demonstrated impressive capabilities in synthesizing photorealistic images from text descriptions.  Yet their effectiveness in precise, grounded image editing remains limited \cite{zhao2024ultraedit, sheynin2024emu}. Unlike text to image generation, editing is an inherently constrained and ill-posed inverse problem that demands fine-grained control over semantics, spatial composition, and alignment with both the input prompt and reference image. 



In this work, we dissect the architectural and training limitations of contemporary generative models for editing and identify three key desiderata essentials: (1) Compositional and Discriminative Visual Representations: The ability to understand scenes as an arrangement of distinct, modifiable local parts rather than a holistic entity, enabling targeted localized edits. (2) Fine-Grained Vision-Language Grounding: A precise alignment between visual features and text to accurately translate textual instructions into visual manipulations. (3) Sufficient Inference-Time Capacity: The computational power to navigate complex solution spaces in challenging inverse problems such as visual editing. 

To achieve these desirable characteristics in diffusion models for editing, we develop new strategies for both training (dual corruption training via masking) and inference (test-time capacity scaling via pause tokens). We hypothesize that current diffusion models, despite their success in generation \cite{rombach2022high}, struggle with precise editing due to limitations in their learned representations. While these models can capture global scene structure well, they lack the granular discriminative features \cite{zhang2023tale} vital for localized manipulation and robust grounding. Features learned via Masked reconstruction \cite{oquab2023dinov2, he2022masked} complement gaussian diffusion with better local understanding \cite{zhang2023tale}. In our work, we combine the benefits of both gaussian diffusion \cite{rombach2022high, xiao2024omnigen} and masked reconstruction \cite{xie2024show, hu2024mask}.

\textit{Dual-Corruption Training via Masking-Augmented Gaussian Diffusion (MAgD)}: 
We develop a novel training method with dual forward process by first adding gaussian noise and then masking out input tokens of noisy-image, which is trained with standard denoised score matching objective. This dual-corruption strategy compels the model to develop highly discriminative, localized features through the imperative of contextual infilling from the masking task, thereby fostering compositional visual representations (Desideratum 1) while retaining the capacity for high-fidelity, globally coherent synthesis from the diffusion process. The resultant representations are further refined by training with rich and descriptive prompts (synthetically generated) resulting in improved fine-grained vision-language grounding (Desideratum 2) aided by discriminative visual representations towards enhanced ability to execute complex, localized edits.

\textit{Inference-Time Capacity Scaling via Pause Tokens}: 
Motivated by inference-time scaling techniques in large language models, such as Chain-of-Thought prompting \cite{wei2022chain} and Scratchpad reasoning \cite{nye2021show}, we introduce a novel mechanism for adaptive inference-time capacity scaling for in-context image generative architecture via Pause Tokens \cite{goyal2023think}. These special tokens embedded within the inference process allow the model to dedicate additional compute to refine its understanding and execute complex edits (Desideratum 3). This effectively boosts model's capacity at inference, enabling it to navigate challenging solution spaces and achieve more precise, coherent, and contextually aware visual manipulations.

In Summary, our work introduces and evaluates novel training and inference techniques addressing key desiderata for visual editing with following contributions:
\begin{itemize}[nosep]
    \item We introduce a dual corruption training that synergistically combines masked image modeling and noise-based denoising. This enables MAgD to learn  discriminative and compositional visual representations, demonstrably enhancing fine-grained vision-language grounding critical for precise editing within computationally and data efficient setting.
    
    \item Inference-Time Scaling with Pause Tokens, a mechanism that enables diffusion based image generative models to \textit{scale in-context capacity at inference}, improving their ability to solve harder, grounded editing tasks (without requiring any retraining).

    \item We empirically establish the effectiveness of MAgD and expressive prompt-based training through comprehensive experiments. Our evaluations span diverse benchmarks, including IdeaBench and Complex-Edit suite featuring expressive prompts, enabling robust assessment of zero-shot editing capabilities. Crucially, our evaluation methodology also captures inherent trade-off between faithfulness and instruction-following within visual editing.
    

\end{itemize}



\vspace{-5pt}
\captionsetup{font=small}
\begin{figure}[htbp] 
    \centering
    \subfloat[\centering MAgD Training \label{fig:magd_loass}]{%
        \includegraphics[trim=220 150 200 15, clip, width=0.48\textwidth, keepaspectratio]{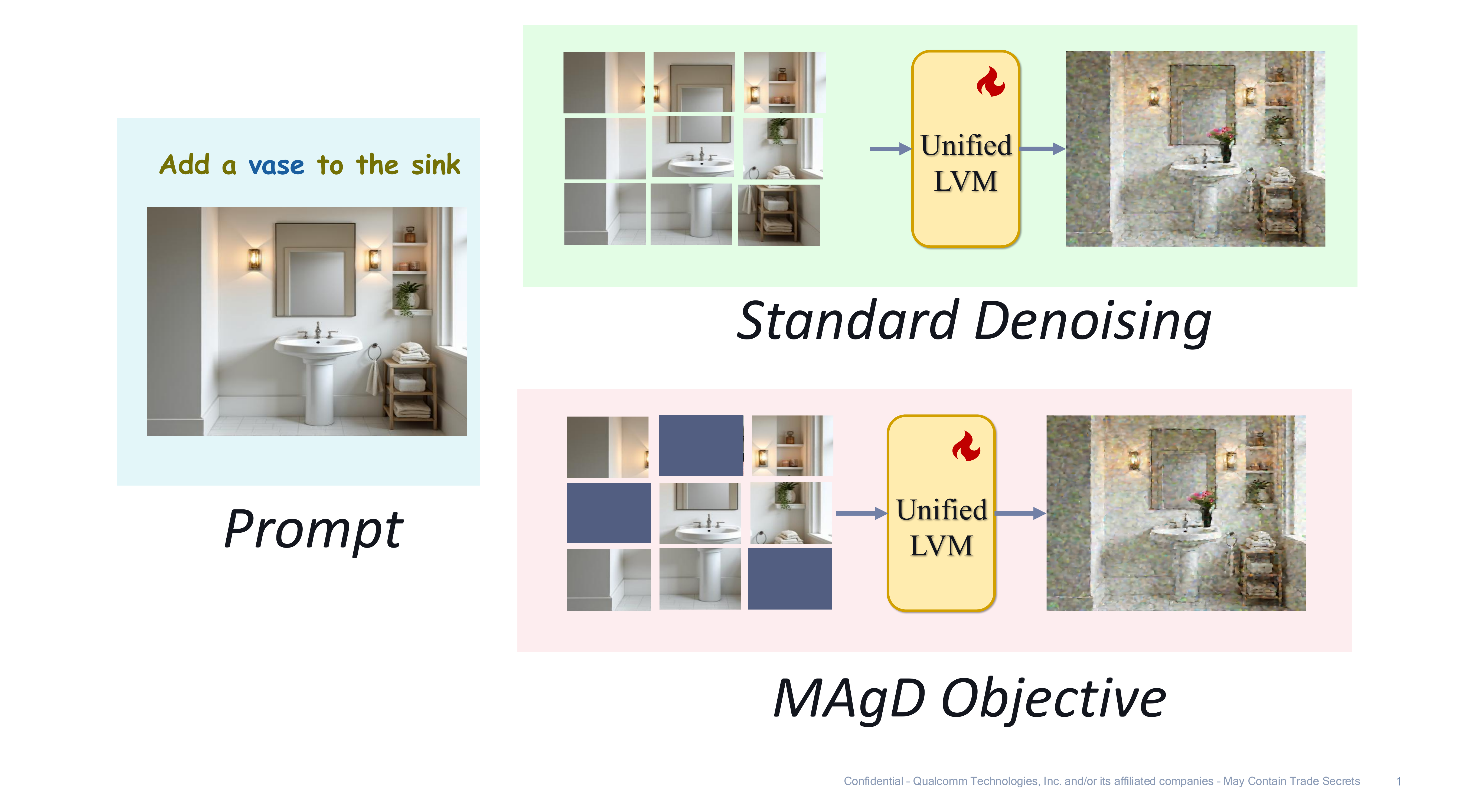}%
    }%
    \hfill 
    \subfloat[\centering Inference Time Scaling\label{fig:inf_scaling}]{%
        \includegraphics[trim=20 50 30 15, clip, width=0.48\textwidth, keepaspectratio]{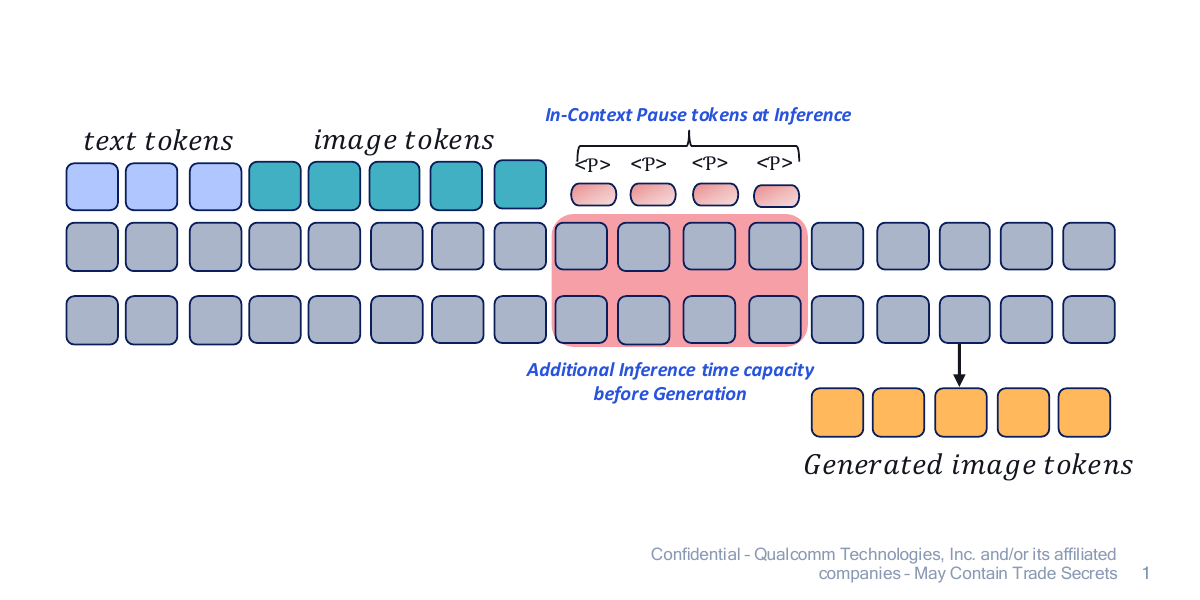}%
    }%
    \caption{a) \textbf {Comparison of Standard Denoising (Top) and MAgD Objective (Bottom)}. The top row illustrates standard denoising, The bottom row depicts the MAgD objective, where the target image is randomly masked, and noise is then added to the entire image before denoising. b) \textbf{Encapsulating Pause tokens for Inference time scaling}: The red box in the figure denotes the impact of additional "thinking time" on latent processing steps  without additional training}
    \label{fig:contribution}%
\end{figure}


\section{Related work}

Masked reconstruction approaches such as MAE \cite{he2022masked} shows strong representation learning via masked reconstruction. While MaskGIT \cite{chang2022maskgit, chang2023muse}  use masked diffusion for synthesis.In contrast, we integrate masked modeling within the continuous diffusion training loop, introducing a dual forward process that better supports partial reconstructions for editing.A closest to our work \cite{zheng2023fast} focusses on training time complexity and not representational properties or editing.

\textit{Editing specific models, inpainting models:} Visual editing, like  "make two people look at each other", is a complex inverse problem requiring identifying regions, modifying them, and maintaining consistency. Standard diffusion models (UNet, DiT) often struggle with the dynamic capacity and context reuse needed for such grounded manipulations. In contrast, in-context architectures inspired by LLMs, such as OmniGen \cite{xiao2024omnigen}, are better suited due to their ability to retrieve and integrate reference information during generation.

\textit{Dense prompt for strong vision-language grounding:} LLM-inspired visual models such a \cite{qi2025cogcom}. \cite{fang2025gotunleashingreasoningcapability}, \cite{xu2025llavacotletvisionlanguage} are emerging as promising frameworks that have shown that integrating dense prompts through CoT style prompting is beneficial for understanding and generation. These architectures treat image generation and editing as a conditioned sequence modeling problem, offering flexible in-context control. Our methods are designed to complement such architectures enhancing their editing capabilities via richer training objectives and inference-time flexibility.


 \begin{figure}[H] 
    \centering
    \includegraphics[width=0.65\textwidth]{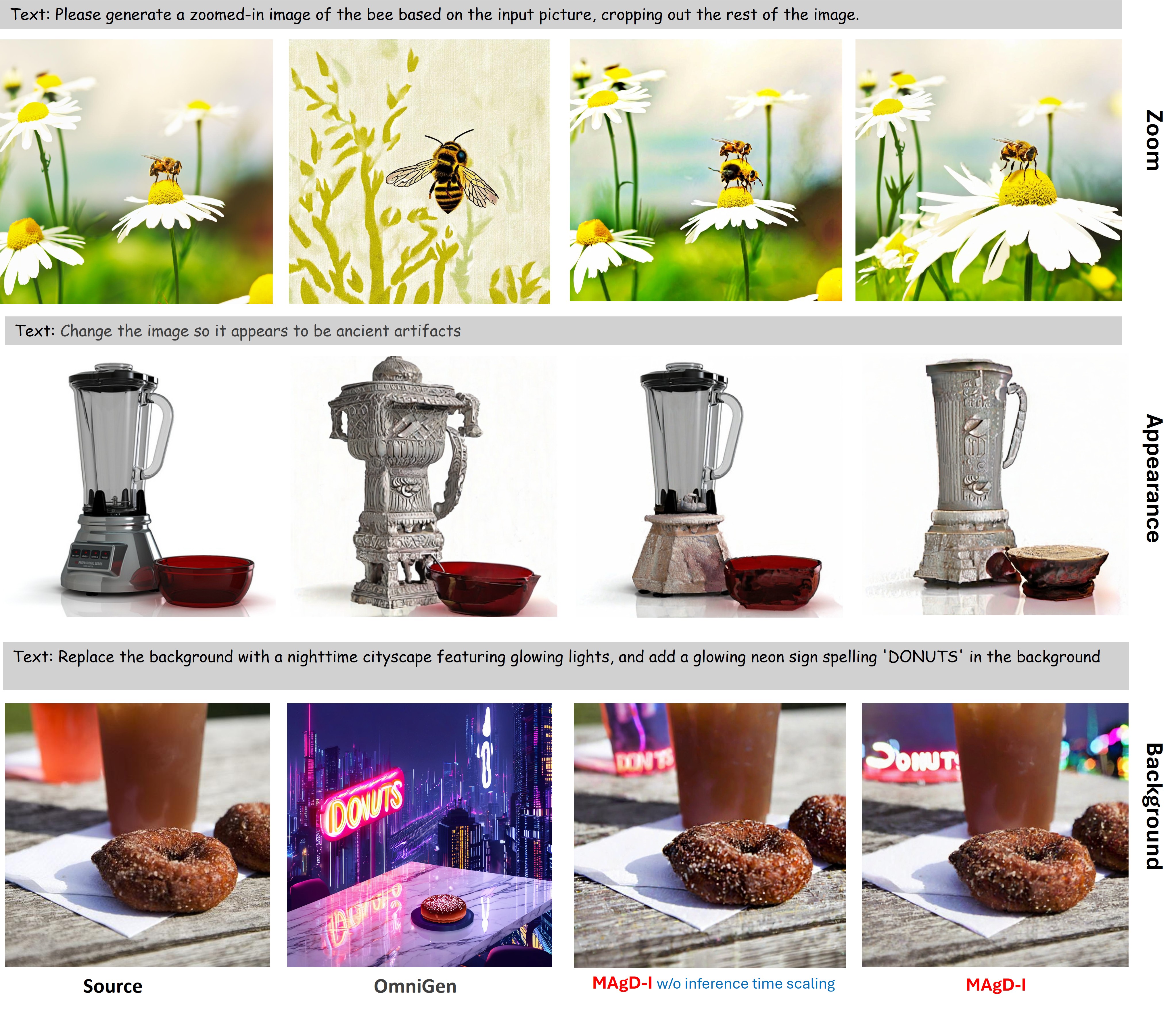}
    \caption{\textbf{Baseline vs MAgD-I (w/o inference scaling) vs MAgD-I}. While baseline follows the prompt, it destroys the scene composition in (top) and (bottom). In (middle) we observe that the object appearance is also modified. While MAgD-I w/o scaling alleviates this problem and retains the scene composition. We observe that with inference scaling, model makes localized updates on the latents more closely following the prompt. In (top) we observe an interesting effect of scaling where the model learns to crop and zoom in on the bee. }
    \label{fig:intro_viz_ideabench}
\end{figure}
\section{Method}
\subsection{Masking-Augmented Gaussian Diffusion}
\label{gen_inst}
We first provide background on masked autoencoders and self-supervised learning, which inspire our proposed masking augmentation to the standard diffusion framework. We then introduce our `\textit{Masking-Augmented Gaussian Diffusion (MAgD)}’ method, a simple yet effective modification to the forward process of diffusion models, to learn effective visual representations for editing.

\subsubsection{Masked Reconstruction for Representational Learning}
Self-supervised learning (SSL) has emerged as a dominant paradigm for representation learning, 
Masked reconstruction adopted by models such as BERT \cite{devlin2019bert} in NLP and MAE \cite{he2022masked} in vision, is a notable SSL strategy. It operates by randomly masking a subset of input tokens or patches and training the model to reconstruct the missing content, thereby encouraging the learning of rich `intra-image' contextual representations. In masked autoencoders, the input $\mathbf{x}$ is partially occluded via a masking operation $\mathcal{M}(\mathbf{x})$, and the model $f_\theta$ is trained to predict the original $\mathbf{x}$ given only the unmasked parts:
\begin{equation}
    \mathbf{\mathcal{L}_{MAE}} = \mathbf{E_{x,\mathcal{M}}} || f_{\theta}(\mathcal{M}(x)) - x ||^2
\end{equation}
where $\mathcal{M}$ is sampled from a distribution over possible masks. By reconstructing the missing information, model builds a holistic, semantically meaningful internal representation of the input. 

\subsubsection{Gaussian Diffusion}
Diffusion models \cite{ho2020denoising, song2020scorebased} learn to reverse a gradual noising process applied to data. A clean sample $\mathbf{x}_0$ is progressively perturbed over $T$ timesteps via a Markovian forward process:
\begin{equation}
    q(x_t|x_{t-1}) = \mathcal{N}(x_t; \sqrt{1-\beta_{t}} x_{t-1}, \beta_{t}\mathbf{I})    
\end{equation}
where $\beta_t$ is a schedule of noise variances. The training objective typically minimizes a simplified denoising score-matching (DSM) loss:
\begin{equation}
    \mathbf{\mathcal{L}_{DSM}}(x_t, \epsilon_\theta) = \mathbf{E_{t,x_0,\epsilon}} || \mathcal{\epsilon}_{\theta}(x_t,t,c) - \epsilon ||^2
\end{equation}
where $\mathbf{\epsilon} \sim \mathcal{N}(0, \mathbf{I})$ is the noise added, $t \in [0,1]$ denotes diffusion and $c$ is conditioning (e.g., text) and $\epsilon_{\theta}$ is a denoising neural network or score function estimator with parameters $\theta$.

\subsubsection{Modified forward Process with Masking}
\label{MFP}
We introduce an auxiliary random masking operation i.e., a secondary corruption within forward process during training. Specifically, given a noisy image $\mathbf{x}_t$, we apply a random binary mask $\mathbf{m} \in {0,1}^d$ sampled independently at each step, with a fixed masking rate $r_{mask}$ (e.g., 0.25). The resulting input from dual forward process to the denoiser is:
\begin{equation} 
    \tilde{\mathbf{x}}_t^{\text{masked}} = \mathbf{m} \odot \mathbf{x}_t \odot \emptyset + (1-\mathbf{m}) \odot \mathbf{x_t}
\end{equation}
where $\odot$ denotes element-wise multiplication and $\emptyset$ represents mask token(e.g., $\emptyset$ is `0' or a learnable mask embedding $e_{\emptyset}$). 
The network $\mathbf{\epsilon}_\theta$ is trained with denoising score matching objective i.e., predict the original noise $\mathbf{\epsilon}$ from the masked and noisy input $\tilde{\mathbf{x}}_t^{\text{masked}}$, given the timestep $t$ and conditioning $c$
\begin{equation}
    \mathbf{\mathcal{L}_{mDSM}}(\tilde{\mathbf{x}}_t^{\text{masked}}, \epsilon_\theta) = \mathbf{E_{t,x_0,\epsilon,\mathbf{m}}} || \mathcal{\epsilon}_{\theta}(\tilde{\mathbf{x}}_t^{\text{masked}},t,c) - \epsilon ||^2
\end{equation}

During MAgD training, the masking-based auxiliary corruption is applied stochastically with a probability 
$p_{magd}$ at each optimization step, following a classifier-free guidance formulation. A uniform random variable 
$u \sim \mathcal{U}(0,1)$ is sampled to determine whether the dual corruption (masking + noise) is applied for a given training instance. Crucially, since our objective is to enhance contextual and compositional representations, we restrict the application of the masking operation to higher noise levels i.e., when the denoising network primarily focuses on low-frequency structural components. This behavior is governed by a time-step threshold parameter $\tau_{MAgD} \in [0,1]$, such that masking is applied only when the diffusion time-step $ t \geq \tau_{MAgD}$ (high noise level). Our overall objective is:
\begin{equation}
\mathcal{L}_{\text{MAgD}} = 
\begin{cases}
   \mathcal{L}_{\text{mDSM}}(\tilde{\mathbf{x}}_t^{\text{masked}}, \epsilon_\theta), & \text{if } u < p_{\text{magd}} \text{ and } t < \tau_{\text{MAgD}} \\
    \mathcal{L}_{\text{DSM}}(x_t, \epsilon_\theta), & \text{otherwise}
\end{cases}
\end{equation}
Our training formulation  enhances latent representations for diffusion models without altering inference-time complexity, which relies on the standard reverse diffusion process. Unlike prior work such as MAE \cite{he2022masked} that performs reconstruction solely in the data space, our method applies supervision across diverse noise levels. 
Our dual corruption serves as both a potent data augmentation and regularization mechanism, compelling the model to learn robust, context-aware representations. Accurately denoising masked regions, particularly under noisy conditions necessitates a sophisticated understanding of spatial semantics and local interactions within the global image context.

We hypothesize that integrating masked reconstruction enhances controllability by improving the model's fidelity to spatial layouts and local structures. This objective promotes stronger contextual reasoning and more discriminative representations, which we posit also improves vision-language alignment. Concurrently, the standard diffusion objective ensures the generation of holistically plausible images consistent with the training distribution. By synergistically combining denoising for global coherence and masked reconstruction for local precision and contextual understanding, our training strategy yields significant benefits for structured and controllable image synthesis.


\subsection{Enhanced Vision-Language Grounding with Expressive Prompts}
Standard text-to-image diffusion models, often trained with simple, holistic prompts (e.g., "a cat sitting on a mat"), struggle with fine-grained compositional understanding and precise, controllable image editing. This limitation is particularly acute in editing tasks, which, by conditioning on both a reference image and a textual instruction, present a more challenging inverse problem than unconditional generation. The target output space is sharply constrained by the input image, demanding accurate interpretation of localized textual commands.





To address this, we advocate for training with expressive prompts. Instead of relying solely on atomic captions, we synthetically generate structured prompts by decomposing complex desired outcomes into a sequence of localized semantic transformations or attribute specifications in sub-prompts $p_i$. This sequence is then concatenated to form a single, richer conditioning signal:
\begin{equation}
    p_{expressive} =  Concat(p_1,p_2,...,p_n)
\end{equation}

Training on such expressive prompts compels the model to internalize fine-grained relationships between textual phrases and their corresponding visual manifestations, thereby significantly enhancing vision-language grounding (VLG) aided by MAgD based improved discriminative visual representations and ability to perform grounded manipulations at inference time.  

Within MAgD training scheme, when an expressive prompt describes attributes or objects intended for a (now masked) region, the model must accurately ground each part of the textual instruction to the corresponding spatial area and semantic content to successfully perform the reconstruction. This process explicitly trains the model to associate fine-grained linguistic elements with specific visual content, even under the noisy and partially occluded conditions inherent to diffusion training.

\subsection{Adaptive Inference-Time Capacity with Pause Tokens for Precise Editing}

Image editing tasks, such as applying a textual instruction to a source image, represent a highly constrained inverse problem. Unlike unconditional generation, the desired output space is significantly restricted by the need to accurately reflect the edit instruction while preserving the unedited aspects of the source image. This precision demand suggests that editing can benefit from mechanisms that allow the model to dedicate more targeted computational effort during inference to resolve complex instructions within this feasible set.

Inspired by approaches that grant models increased "thinking time" \cite{wei2022chain, goyal2023think, nye2021show}, we introduce Pause Tokens $\langle \texttt{pause} \rangle$ special, predefined tokens inserted into the textual prompt exclusively at inference time. 
Formally, at inference, we modify the input prompt by inserting pause tokens $\langle \texttt{pause} \rangle$ after given instruction, reference image before image generation.
\begin{equation}
 p_{paused} = Concat(p_1, ref_{img}, \langle \texttt{pause} \rangle, \langle \texttt{pause} \rangle, \dots, \langle \texttt{pause} \rangle  )   
\end{equation}
where $\langle \texttt{pause} \rangle$ is a learnable or predefined placeholder token indicating processing boundaries and $p_1$ is given instruction prompt, $ref_{img}$ is given reference or source image.

Key aspects of our pause token scaling strategy are, \textit{Inference-only Application}: This strategy is inference-only, decoupling the need for increased capacity from the training phase.

\textit{Enhanced Contextual Integration}: The additional processing steps induced by pause tokens can facilitate a better integration of the text instruction with the visual features, crucial for precise localization. Pause tokens can function as soft segmentation markers within the prompt, enabling the model to leverage additional computation to different conceptual parts of the editing task. For instance, a pause might help to delineate region to be preserved from modification (e.g., add or remove, etc.) from its surrounding context. This provides the model with implicit "processing boundaries," allowing it to better internalize and ground distinct operational aspects of a edit instruction which involves complex sub-edits before generating the final output.

\textit{Improved Grounded Editing}: Empirically, pause tokens improve the model's ability to maintain faithfulness to reference images and apply fine-grained edits. We observe better improvements when the model is trained with expressive prompting, which encourages the learning of enhanced representations and effective attention-based mechanisms capable of utilizing the additional computational capacity introduced at inference. This highlights a path towards dynamically allocating effective computational resources at inference, guided by the structure of the task itself.


\vspace{-5pt}
\section{Experiments}
\vspace{-5pt}
\paragraph{Training Data} 
We curate a comprehensive training dataset of approximately 450K samples, evenly split between text-to-image (T2I) generation and image editing, to facilitate the fine-tuning of OmniGen.\cite{Omnigengit} This dataset draws from high-quality sources including UltraEdit (250K samples) \cite{zhao2024ultraedit}, MagicBrush \cite{zhang2023magicbrush}, and resources introduced in Aurora \cite{krojer2024learningactionreasoningcentricimage} (specifically KubricEdit and Something-Something-Edit). Beyond standard (image, instruction) pairs, we incorporate richer (image, mask, instruction) triplets from MagicBrush and UltraEdit and include diverse local and compositional editing examples to broaden coverage. Furthermore, for controlled experiments on semantic guidance, we augment a subset to create a specialized corpus featuring Expressive Edit Plan Annotations—step-wise plans generated by MLLMs for (image, instruction) pairs (details in Appendix). This controlled corpus includes 100K samples with the full \textit{(image, instruction, edit plan)} triplets and 100K with only \textit{(image, instruction)} pairs, allowing us to study the impact of step-wise training.

\vspace{-7pt}
\paragraph{Metrics}
We employ both standard quantitative metrics and MLLM-based evaluations to holistically assess model performance. \textit{CLIP-Dir} is our primary directional metric, capturing alignment between the edit instruction and the semantic shift between the input and output image. \textit{CLIP-Img} and \textit{DINO} measure faithfulness to the source image. \textit{CLIP-T} evaluates textual alignment with the edit instruction, but is less sensitive to structural artifacts or visual realism.
We find that CLIP-Dir vs CLIP-Img tradeoffs for evaluating edit quality and offer an informative \textit{2D frontier} to assess edit quality (details in Appendix). Our goal is to maximize directional accuracy while minimizing unnecessary deviation from the input image.
\newline
\textbf{MLLM-Based Evaluation}: As shown in prior work \cite{wei2024omniedit},\cite{yang2025textttcomplexeditcotlikeinstructiongeneration}, standard metrics often poorly correlate with human preferences. To address this, we adopt MLLM-based evaluation across four benchmarks using the following dimensions:
Instruction Faithfulness (IF), Identity Preservation(IDP), Perceptual Quality (PQ). We report MLLM(Aggregate) as a unified metric.This score is weighted to de-emphasize scenarios where the model successfully executes instructions but introduces unnecessary scene composition refactoring.

\vspace{-7pt}
\paragraph{Image Editing Evaluation Benchmarks}
We evaluate on several established test sets: Emu-Edit \cite{sheynin2024emu}, MagicBrush\cite{zhang2023magicbrush}, and the recently proposed Complex-Edit benchmark \cite{yang2025textttcomplexeditcotlikeinstructiongeneration}, which assesses robustness under multi-step and compositional edit instructions. To gauge generalization to real-world T2I tasks, we also test on IdeaBench \cite{liang2024idea}, which includes image enlargements, retouching, style transformations, and branded content generation. More details on this will be discussed in Results.


\vspace{-7pt}
\paragraph{Finetuning Setup} 
We build upon OmniGen \cite{xiao2024omnigen} as our base architecture due to its unified token space for image and text and its versatility across modalities. OmniGen is pretrained across diverse tasks with a diffusion score-matching objective, and we finetune it for our MAgD framework using the same objective and curriculum. This allows us to isolate the benefits of our proposed augmentation strategy.

\vspace{-7pt}
\paragraph{Training Setup}
We train our model on lr=1$\exp{-4}$ with a warm up schedule followed by cosine scheduling post warm up stage. We train on a batch size of 128. The model is trained approximately on ~4000 gradient steps. We train on 4 A100 GPUs. We follow the Omnigen prompting strategy of interleaving multi-modal prompts. We condition on masked source image whenever masks are available in dataset. We follow a similar CFG as baseline with $ 10\% $ of prompts being empty. All our training samples are of resolution 1024.
During inference we do not condition on masked images we use a guidance scale of 2.5 and image guidance of 1.5.

\vspace{-7pt}
\paragraph{Baselines}: We compare against MagicBrush \cite{zhang2023magicbrush}, Emu-Edit \cite{sheynin2024emu}, and InstructPix2Pix \cite{brooks2023instructpix2pixlearningfollowimage}, An alignment focused model Seed-Edit \cite{shi2024seededitalignimageregeneration} and on Show-O \cite{nye2021show} also an MLLM trained on masked diffusion objective for generation and understanding. 


\subsection{Image Editing Benchmarking} 
\vspace{-5pt}
We empirically evaluate our core contributions: the Masking-Augmented Gaussian Diffusion (MAgD) objective, Expressive Prompt finetuning, and adaptive inference-time scaling. We analyze their impact on instruction adherence, source image faithfulness, and overall edit quality, contextualizing findings within the inherent trade-offs of image editing and highlighting improvements in vision-language grounding.

\begin{table}[tbp]
\centering
\begin{threeparttable}
\small 
\caption{This table compares the behavior of \textbf{Omnigen} and \textcolor{blue}{MAgD-I} and other SoTA benchmarks . While Omnigen follows instructions, it exhibits a tendency to modify regions beyond the specified edits in the source image. \textcolor{blue}{MAgD} addresses this by improving on \textbf{perceptual similarity (DINO, CLIP-I) }with source image while maintaining the baselines directional similarity.  By employing  inference scaling \textcolor{blue}{MAgD} improve further  \textbf{balances instruction following and source preservation.}}
\label{tab:editing_results_v2}
\renewcommand{\arraystretch}{1.1} 
\setlength{\tabcolsep}{2pt} 
\begin{tabular}{@{}
  l 
  c 
  S[table-format=1.3, table-parse-only, table-text-alignment=center] 
  S[table-format=1.3, table-parse-only, table-text-alignment=center] 
  S[table-format=1.3, table-parse-only, table-text-alignment=center] 
  S[table-format=1.3, table-parse-only, table-text-alignment=center] 
  S[table-format=1.1, table-parse-only, table-text-alignment=center] 
  @{}
}
\toprule
\rowcolor{headergray}
\textbf{Model} & \stackth{Train Data} & \stackth{CLIP-I ($\uparrow$)} & \stackth{DINO ($\uparrow$)} & \stackth{CLIP-T ($\uparrow$)} & \stackth{CLIP-Dir ($\uparrow$)} & \stackth{MLLM ($\uparrow$)} \\
\midrule
\rowcolor{sectiongray} 
\multicolumn{7}{@{}l@{}}{\textbf{Emu-Edit Benchmarking }} \\ \addlinespace[0.5ex]
InstructPix2Pix \cite{brooks2023instructpix2pixlearningfollowimage} & {--} & 0.852 & 0.766 & 0.274 & 0.078 & {--} \\
MagicBrush & {--} & 0.918 & 0.892 & 0.276 & 0.066 & {--} \\
EmuEdit & {--} & 0.862 & 0.836 & 0.284 & 0.107 & {--} \\
UltraEdit & {--} & 0.845 & 0.794 & 0.283 & 0.108 & {--} \\
SeedEdit (SDXL) & {--}   & 0.803 & \multicolumn{1}{c}{N/A} & \multicolumn{1}{c}{N/A} & 0.116 & {--} \\
\cdashline{1-7}
Omnigen  & {N/A} & 0.820 & 0.882 & 0.264 & 0.122 & 8.00 \\
\rowcolor{lightgray}
\rowcolor{lightgray}
MAgD-I & {400k} & {0.869} & {0.894} & {0.263} & \textbf{0.134} & \textbf{8.43}  \\ 
\rowcolor{lightgray}
\hspace{0.5cm} w/o Inference Scaling & {400k} & \textbf{0.873} & \textbf{0.927} & \textbf{0.269} & 0.126 & {8.40}\\

\midrule
\rowcolor{sectiongray} 
\multicolumn{7}{@{}l@{}}{\textbf{Magicbrush Benchmarking }} \\ \addlinespace[0.5ex]

Show-O (512) & {--} & {0.61} & 0.630 & 0.160 & 0.045 &  {--}  \\
Omnigen (512)   & {--} & {0.649} & 0.686 & 0.159 & 0.048 & {--} \\ 
EmuEdit (1024) & {--} & 0.897 & 0.879 & 0.261 & {--} & {--} \\
Omnigen (1024)  & {--} & 0.854 & 0.904 & 0.302 & 0.050 & 7.58 \\
\rowcolor{lightgray}
MAgD-I (1024) & {400k} &  0.834  & 0.902 & 0.308 & \textbf{0.063} & \textbf{7.91} \\
\rowcolor{lightgray}
\hspace{0.5cm} w/o Inference Scaling (1024) & {400k} & \textbf{0.875} & \textbf{0.932} & \textbf{0.311} & 0.059 & 7.89 \\ 
\bottomrule
\end{tabular}
\begin{tablenotes}[flushleft]
\footnotesize 
\item \textbf{Notes:} . 
All values are reported as performance scores (higher is better $\uparrow$). `{--}` denotes missing or unreported values. `{N/A}` denotes Not Applicable. CLIP-I, DINO, and CLIP-T refer to different perceptual similarity. metrics. MLLM reported here is weighted average across Instruction following (IF) and Identity preservation ( IDP). 
\end{tablenotes}
\end{threeparttable}
\end{table}

\begin{table}[tbp]
\scriptsize
\renewcommand{\arraystretch}{1.6}
\setlength{\tabcolsep}{3pt}
\makebox[\textwidth][c]{ 
\begin{minipage}[t]{0.5\textwidth}
\centering
\begin{tabular}{lcccc}
\toprule
\rowcolor{headergray}
\textbf{Method} & \textbf{Avg} & \textbf{IFP} & \textbf{IDP} & \textbf{PQ} \\
\midrule
Omnigen & 7.53 & 6.42 & 6.39 & 7.15 \\
MAgD         & 8.62 & 6.79 & 7.67 & 6.48 \\
\bottomrule
\end{tabular}
\vspace{0.4em}
\subcaption{\textbf{Complex Edit Benchmark}:  MAgD vs Omnigen across different prompt augmentation complexities. Where \textbf{IFP}, \textbf{IDP} and \textbf{PQ} follow the protocol of \cite{yang2025textttcomplexeditcotlikeinstructiongeneration}. With masked objective, IDP flattens with increasing complexity allowing the model to complete multiple  edits without interfering with each other. As evidenced in \ref{fig:MaGD on complex edit}}
\label{tab:complex_edit}
\end{minipage}
\hspace{0.8em} 
\begin{minipage}[t]{0.5\textwidth}
\centering
\begin{tabular}{lc}
\toprule
\rowcolor{headergray}
\textbf{Method} & \textbf{MLLM (Aggregate)} \\
\midrule
Omnigen & 57.41\\
MAgD & 74.07\\ 
\bottomrule
\end{tabular}
\vspace{0.4em}
\subcaption{\textbf{Idea Bench Success Rate} MAgD vs Omnigen on I2I tasks only. We evaluate across all I2I tasks except on couple icon and id photo generation. Interestingly we observe greater improvements on Enlargement, attribute editing and brand merchandise editing. In these tasks Omnigen fails to retain the source image or the background in case of attribute editing. More details in appendix. We provide a sample in \ref{fig:intro_viz_ideabench}}
\label{tab:idea_bench}
\end{minipage}
}
\vspace{-20pt}
\end{table}

\setlength{\belowcaptionskip}{0pt}
\vspace{-17pt}
\begin{figure}[H] 
    \centering
    \includegraphics[width=1\textwidth]{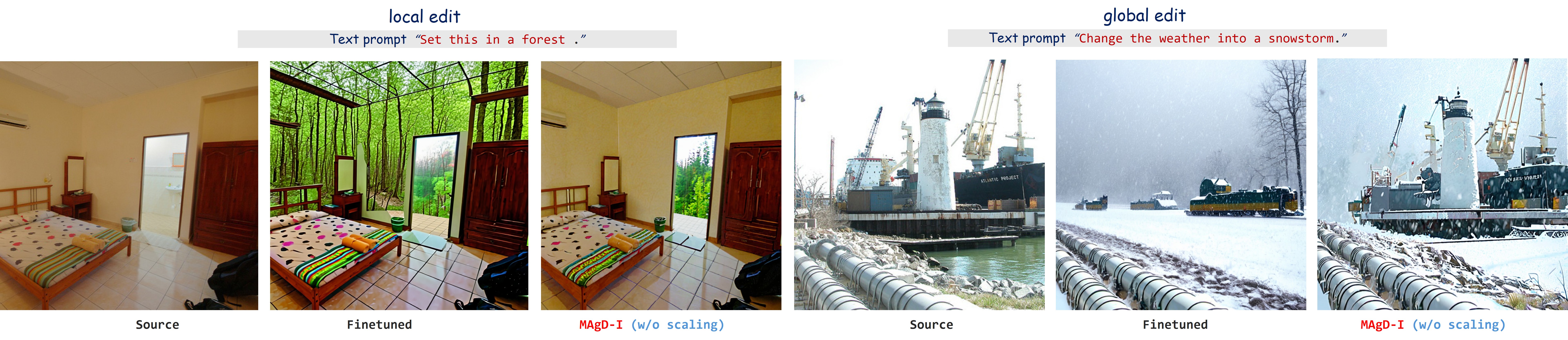}
    \vspace{-15pt}
    \caption{\textbf{Comparisons between finetuned-Omnigen vs MAgD w/o inference scaling} across different tasks including \textit{local}, \textit{global}, \textit{background}, \textit{remove}. When using our objective the model retains the intra image composition on global edit task while changing the weather. While finetuned omnigen loses the structure. In the local edit, MAGD improves the semantic representations for example, MAGD generates a good composition and semantically meaningfull image of bedroom inside a forest. }
    \vspace{-30pt}
    \label{fig:MaGD_effectivenss}
\end{figure}

\vspace{-10pt}
\paragraph{MAgD Objective:} Significantly enhances editing performance. As shown in Table~\ref{tab:editing_results_v2}, MAgD substantially improves both source image faithfulness (DINO:$0.882$ to $0.927$) and instruction adherence (CLIP-Dir: $0.122 \to 0.126$) over OmniGen baseline. This simultaneous advancement in typically competing metrics indicates that MAgD fosters a more robust vision-language grounding, enabling the model to better reconcile edit instructions with image content. 

\begin{figure}[H] 
    \centering
    \includegraphics[height=7cm, width=0.9\textwidth]{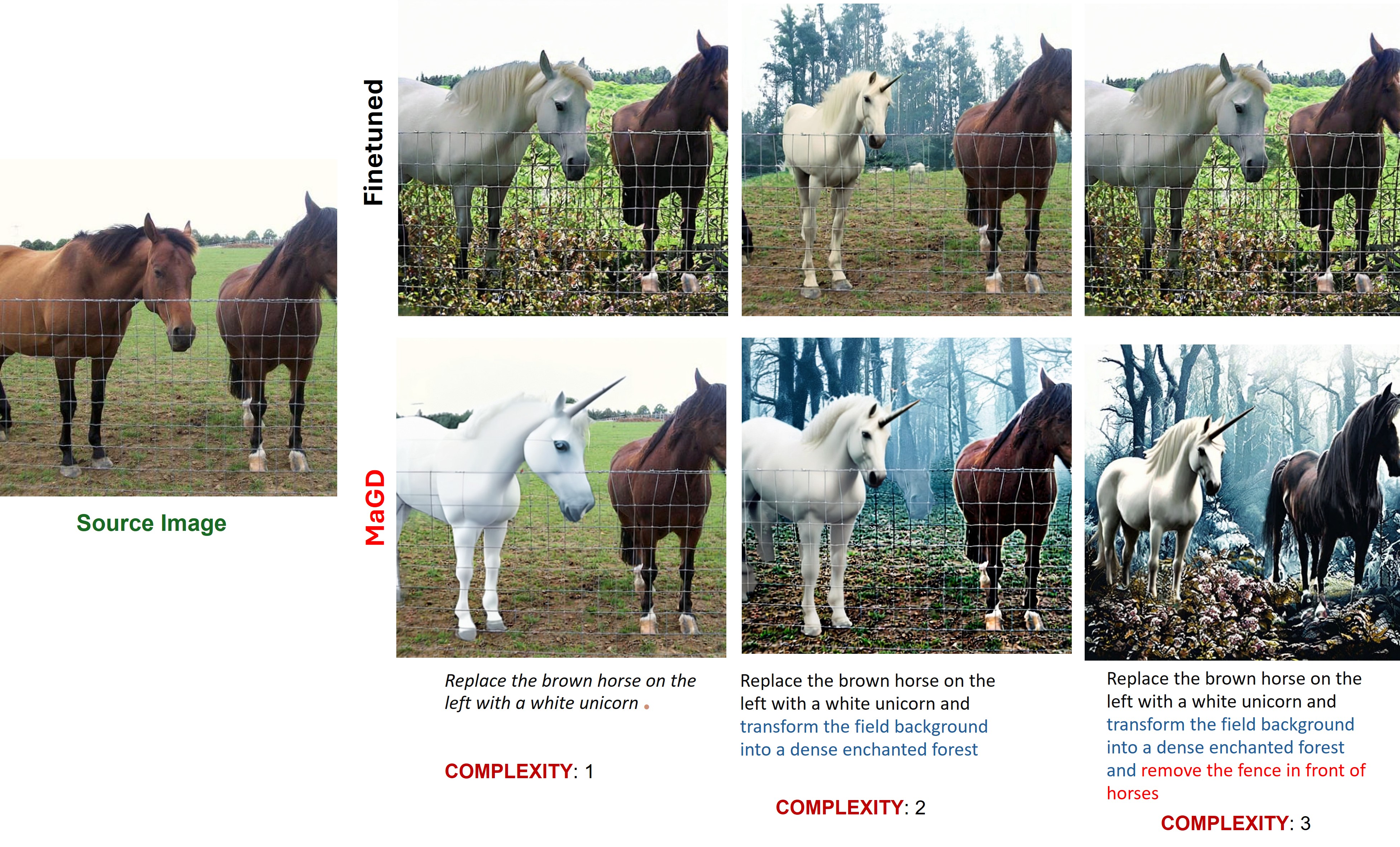}
    \caption{[Top]:\textbf{Finetuned model} vs [bottom] \textbf{\textcolor{red}{MAgD-I}} for increasing complexities of same instruction prompt and source image. On complexity 1, the finetuned model changes the color of horse but unicorn is clearly generated for all complexities for MAgD-I. At higher complexities (Complexity 2 and 3), MAgD exhibits strong semantic understanding and contextualization, successfully replacing the background with an enchanted forest and precisely removing the fence (Complexity 3). Notably, MAgD maintains the intra-image composition throughout these complex edits, a key advantage over the finetuned baseline which introduces significant structural artifacts and fails to follow the full instructions.}
    \label{fig:MaGD on complex edit}
\end{figure}

These benefits generalize across benchmarks. MAgD consistently outperforms strong baselines like SeedEdit (SDXL) and UltraEdit, alongside OmniGen, on both traditional metrics (Table~\ref{tab:editing_results_v2}). Specifically, on complex benchmarks like Complex-Edit and IdeaBench (Table~\ref{tab:complex_edit} \ref{tab:idea_bench}), MAgD excels in instruction following (IFP) while preserving source image structure (IDP). This strong performance on nuanced edits further evidences MAgD's enhanced grounding capabilities. Qualitative examples (Figures~\ref{fig:MaGD_effectivenss}, \ref{fig:MaGD on complex edit}, \ref{fig:intro_viz_ideabench}) visually corroborate MAgD's ability to execute semantically precise edits, maintaining contextual coherence a direct outcome of improved vision-language understanding and improved compositional representations.

Crucially, an ablation study (Table~\ref{tab:ablation_study}) confirms MAgD's intrinsic contribution: even without Expressive Prompt finetuning, MAgD alone boosts both CLIP-Dir and DINO over OmniGen and finetuned omnigen. This isolates the objective's direct impact on enhancing the model's foundational vision-language alignment.

\begin{table*}[tbp] 
  \centering
  \caption{Ablation study on dense prompt design choices and their impact on performance metrics.
           \greencheck indicates the component is included, \redx indicates it is excluded.
           Metrics: CLIP-I (Image), DINO, CLIP-T (Text), CLIP-DIR (Directional), MLLM, IFP, IDP.  }
  \label{tab:ablation_study}
  \small 
  \begin{adjustbox}{width=\textwidth}
  \begin{tabular}{@{}*{3}{c}ccccccc@{}} 
    \toprule
    \rowcolor{headergray}
    FineTuning & DP & MAgD & CLIP-I & DINO & CLIP-T & CLIP-DIR & $MLLM_{avg}$ & IFP & IDP \\
    \midrule
    \redx       & \redx       & \redx          & 0.820 & 0.882 & 0.264 & 0.122 & 8.00 & 7.40 & 8.30 \\
    \greencheck & \redx       & \redx            &    0.862   &  0.917      & 0.264      & 0.111      & 8.10     &  7.44    &  8.50     \\
    \greencheck & \greencheck & \redx             & 0.862 & 0.916 & 0.266 & 0.115 &  7.95    & 7.15     & 8.54     \\
    \rowcolor{sectiongray} 
    \multicolumn{10}{@{}l@{}}{\textbf{MADI  Training} i.e., masking augmented dual corruption forward process } \\ \addlinespace[0.5ex]
    \greencheck & \redx       & \greencheck        & 0.849 & 0.913 & 0.268 & 0.127 & 8.30 & 7.70 & 8.60 \\
    \greencheck & \greencheck & \greencheck       & 0.873 & \textbf{0.927} & 0.269 & 0.126 & 8.40  & 7.81 & 8.79  \\
    \bottomrule
  \end{tabular}
  \end{adjustbox}
\end{table*}

\vspace{-10pt}
\paragraph{Expressive Prompt Finetuning}
Finetuning with Expressive Prompts (EP) further refines editing capabilities, synergizing with the MAgD objective. Our ablations (Table~\ref{tab:ablation_study}) demonstrate distinct benefits: For the baseline OmniGen, EP finetuning primarily elevates instruction following (CLIP-DIR:$0.110 \to 0.115$), likely by exposing the model to more explicit, decomposed edit narratives that improve its interpretation of textual commands.

When applied to MAgD, which already possesses strong instruction adherence due to its improved grounding, EP finetuning yields further gains in faithfulness (DINO) and edit precision. This suggests a complementary effect: MAgD establishes a robust, well-grounded representational foundation, which EPs then leverage to refine the model's interpretation and execution of complex, localized instructions. These combined improvements are consistently reflected in MLLM benchmark scores (e.g., Tables~\ref{tab:ablation_study}, \ref{tab:complex_edit}, \ref{tab:idea_bench}), underscoring the broad impact on edit quality and the model's capacity for sophisticated vision-language reasoning.

\vspace{-10pt}
\paragraph{Inference time scaling}
Our proposed inference-time scaling strategy, using "pause tokens," offers a new degree of freedom to modulate the balance between instruction adherence and image faithfulness in editing. This is critical as editing inherently involves a trade-off between these two aspects \cite{shi2024seededitalignimageregeneration}. Our method allows navigating this trade-off without retraining. For instance, with MAgD, scaling improved CLIP-Dir (instruction adherence) from $0.126$ to $0.134$ with a negligible DINO (faithfulness) decrease from ($0.873 \to 0.869$) \ref{tab:editing_results_v2}.

By adjusting number of pause tokens, practitioners can explore the Pareto frontier of this trade-off (Table~\ref{tab:inference_scaling_1}). For MAgD, optimizing for faithfulness increased DINO from $0.927 \to 0.943$
. For Omnigen, faithfulness optimization yielded DINO $0.882 \to 0.919$(at the cost of CLIP-Dir $0.122 \to 0.092$), while adherence optimization improved CLIP-Dir from $0.122 \to 0.139$.

Crucially, inference-time scaling enhances instruction adherence even at a fixed target faithfulness (Table~\ref{tab:inference_scaling_1}). For MAgD, at a DINO threshold of 0.91,CLIP-Dir improved ($0.101 \to 0.110$), and importantly, the recall (percentage of edits meeting the DINO target) substantially increased ($0.73 \to 0.85$). Omnigen showed similar gains: at a 0.91 DINO target, recall rose ($0.75 \to 0.82$) alongside CLIP-DIR ($0.090\to 0.100$). Similar benefits arise when targeting a specific CLIP-DIR.

This inference-time controllability, to our knowledge a first for in-context multi-modal editing architectures, allows users to select optimal operating points based on application needs. The approach is potentially generalizable, and future work could explore richer feedback, especially for constrained inverse problems.


\begin{table}[t]
\centering
\small
\renewcommand{\arraystretch}{0.95}
\setlength{\tabcolsep}{2.8pt}
\resizebox{\textwidth}{!}{%
\begin{tabular}{
  l !{\vrule} 
  S[table-format=1.3] S[table-format=1.3] !{\vrule} 
  S[table-format=1.3] S[table-format=1.3] !{\vrule} 
  S[table-format=1.3] S[table-format=1.2] S[table-format=1.2] !{\vrule} 
  S[table-format=1.3] S[table-format=1.3] S[table-format=1.2]
}
\toprule
\rowcolor{headergray}
\textbf{Model}
& \multicolumn{2}{c!{\vrule}}{\textbf{Best CLIP}} 
& \multicolumn{2}{c!{\vrule}}{\textbf{Best DINO}} 
& \multicolumn{3}{c!{\vrule}}{\textbf{DINO $\geq$ 0.91}} 
& \multicolumn{3}{c}{\textbf{CLIP-DIR $\geq$ 0.09}} \\
\cmidrule{2-3} \cmidrule{4-5} \cmidrule{6-8} \cmidrule{9-11}
 & {DINO} & {CLIP-DIR} & {DINO} & {CLIP-DIR} & {DINO} & {CLIP-DIR} & {Recall} & {DINO} & {CLIP-DIR} & {Recall} \\
\midrule
Finetuned (w/ Expressive prompts)                     & {N/A}  & {N/A}  & {N/A}  & {N/A}  & 0.969 & 0.090 & 0.75  & 0.895 & 0.221 & 0.44 \\
\hspace{0.5cm} w Inference Scaling    & 0.864  & 0.139  & 0.919  & 0.092  & 0.961 & 0.110 & 0.82  & 0.887 & 0.190 & 0.57 \\
MAgD  & {N/A} & {N/A} & {N/A} & {N/A} & 0.970 & 0.101&0.73 &0.900 & 0.230 &0.47 \\
\hspace{0.5cm} w Inference Scaling  & 0.894  & 0.134  & 0.943  & 0.087  & 0.962 & 0.110 & 0.85  & 0.912 & 0.195 & 0.54 \\
\bottomrule
\end{tabular}%
}
\caption{\textbf{Comparison of MAgD and Finetuned Omnigen with Expressive Prompts under different scoring and recall settings}. Evaluated on CLIP-DIR and DINO thresholds. Evaluation metrics resulting from different think token selection criteria per sample. Columns 1 and 2 show average CLIP-DIR and DINO scores when tokens are selected by maximizing individual CLIP and CLIP-DIR, respectively. Columns 3 and 4 report Recall metrics for a specific target (@ Target DINO and @ Target CLIP), indicating the rate at which a token meeting the specified target score can be found. }
\label{tab:inference_scaling_1}
\end{table}


\setlength{\belowcaptionskip}{0pt}


\subsection{Benchmarking T2I generation}
To further assess the efficacy of MAgD in enhancing semantic understanding and compositional generalization in text-to-image models, we conduct a rigorous evaluation on the GenEval benchmark. GenEval is specifically designed to probe a model's ability to handle challenging prompts involving multiple objects, attributes, spatial relationships, and counting, thereby providing a more nuanced understanding of generation capabilities beyond standard FID/IS scores.

We also evaluate MAgD on GenEval \cite{ghosh2023genevalobjectfocusedframeworkevaluating} across different sub-tasks as seen in \ref{tab:performance_metrics_scores_only}. We compare a Baseline model fine-tuned i.e Base-FT in the same training setup against our proposed Composite model (utilizing the MAgD objective). We observe that vanilla finetuning on the OmniGen objective can deteriorate the performance. While for the same training sample size when finetuned with masked objective we notice a bigger improvement especially in two object scenarios and color attribute scenarios. 

\begin{figure}[H] 
    \centering
    \includegraphics[height=7cm, width=0.9\textwidth]{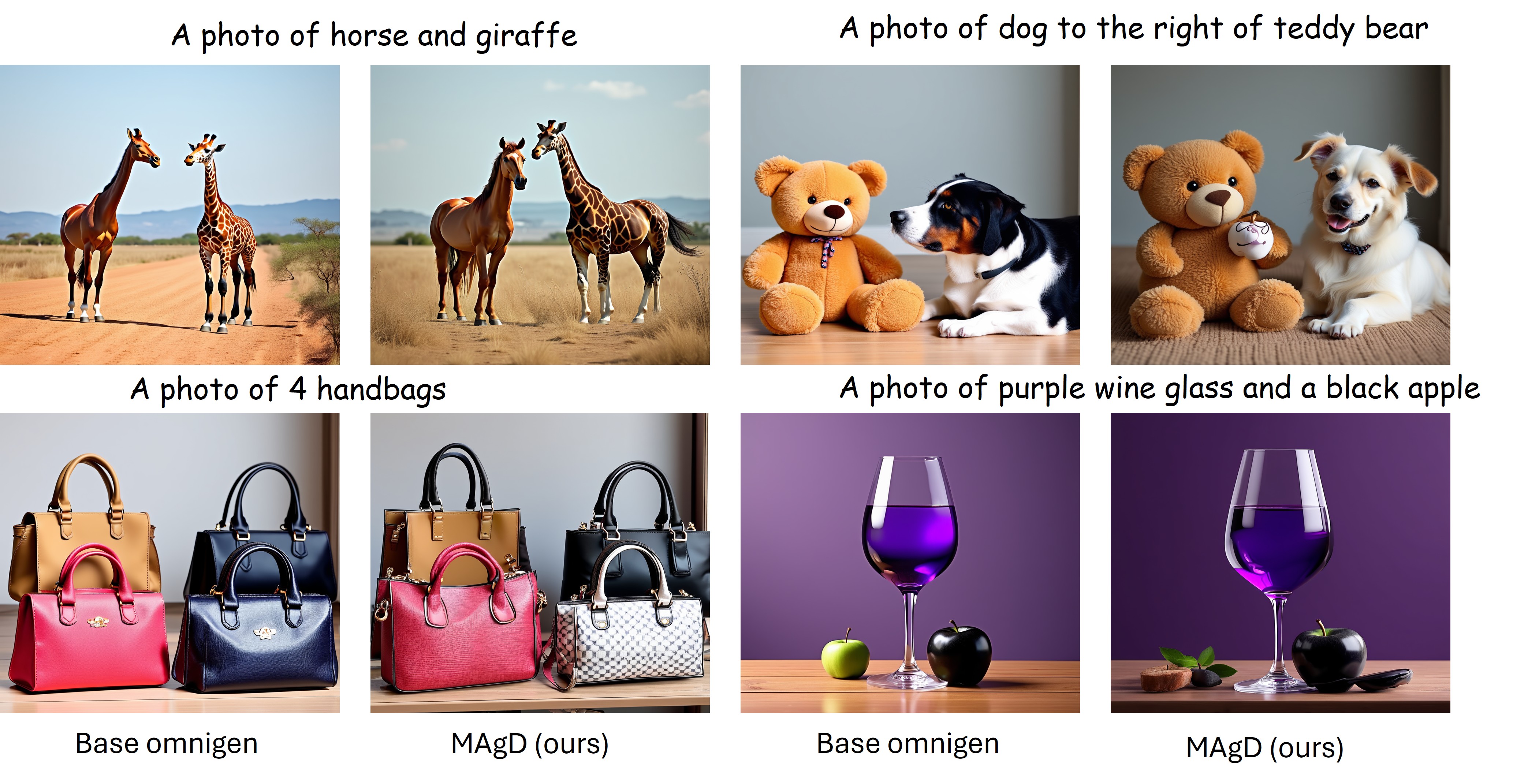}
    \caption{\textbf{Finetuned model} vs  \textbf{\textcolor{red}{MAgD}} We show the Qualitative results on count and two object tasks on GenEval. In the first figure, sometimes Base model fails to generate two distinct objects as evidenced in the first figure of "A photo of a horse and giraffe" }
    \label{fig: GenEval}
\end{figure}

The core hypothesis underpinning MAgD is that its dual-corruption strategy, particularly the masking objective at high noise levels, encourages the model to learn more robust and disentangled semantic representations. Such representations are crucial for accurately interpreting and synthesizing images from complex textual descriptions that require precise compositional reasoning. GenEval, with its diverse set of prompts targeting these compositional skills, serves as an ideal testbed to validate this hypothesis and demonstrate MAgD's practical benefits in scene composition. Please find some example generations in \ref{fig: GenEval}

\begin{table}[htbp]
    \centering
    \caption{\textbf{Performance Metrics on Geneval}. Results are demonstrated by averaging across fixed 5 seeds. The seeds Omnigen used are not released hence these results are different from that in the paper.  
}
    \label{tab:performance_metrics_scores_only}
    \begin{tabular}{l *{7}{c}} 
        \toprule
        & Position & Colors & Color\_attr & Counting & Single object & Two object & Avg \\
        \cmidrule(lr){2-2} \cmidrule(lr){3-3} \cmidrule(lr){4-4} \cmidrule(lr){5-5} \cmidrule(lr){6-6} \cmidrule(lr){7-7} \cmidrule(lr){8-8}
        \addlinespace[0.5ex] 
        \midrule
        Base-FT & 24.75 & 81.91 & 41.25 & 49.69 & 97.81 & 66.67 & 60.30 \\
        Ours & 27.5 & 82.71 & 45.25 & 49.69 & 99.06 & 76.77 & 63.49 \\
        \bottomrule
    \end{tabular}
\end{table}

\section{Limitations and Future Work}
This work demonstrates the efficacy of MAgD, our novel dual corruption process, and inference-time scaling within a specific in-context architecture, finetuned at a modest scale for image generation/editing. Consequently, the performance benefits of scaling our approach with larger models, more extensive datasets, and to other modalities (beyond vision, despite LLM initialization) are yet to be fully explored. Future research could focus on enhancing the synergy between diffusion processes and next-token prediction and in-context models towards versatile multi-modal models.

\clearpage 
{
    \small
    \bibliographystyle{plain}
    \bibliography{main}
}

\clearpage

\section{Supplementary Contents}
\label{sec:SuppleIntro}

As part of the supplementary materials for this paper, we share our Implementation details, show extended qualitative and quantitative results and provide additional theoretical analysis for our proposed approach. The supplementary materials contain: 
\begin{easylist}[itemize]
@ Inference time scaling
    @@ Effect of Inference time scaling, additional evaluations
    @@ Trade-offs: Maximizing Instruction Following under Faithfulness Constraints
    @@ Comparative Analysis on Training objectives.
    @@ Evaluation on complex prompts (Complex Edit)
    @@ Future Directions for Inference Capacity Scaling:
@ MAgD Design Choices Ablation
    @@ Noise level Threshold Selection
    @@ Ablations on Masking rate
    @@ Efficacy of learnable mask
@ Limitations of exisiting metrics
    @@ CLIP-T visualizations
    @@ Trade-off CLIP-T vs CLIP-Dir
    @@ CLIP-DIR vs Instruction following
    @@ Analysis of CLIP scores on spatial relationships and attribute editing
@ Experimental Setup
    @@  Training datasets 
    @@  Expressive prompt corpus
        @@@  Example expressive prompts
    @@  Evaluation Benchmarks
    @@  Hyper parameters 
@ Qualitative evaluations

\end{easylist}


\subsection{Harnessing Inference-Time Capacity Scaling via Pause Tokens for Visual Editing}
A central challenge in diffusion-based visual editing is the inherent capacity limitation of standard pipelines when confronted with complex instructions or inverse problems. This work investigates the hypothesis that dynamically increasing effective network capacity at inference time can significantly improve editing performance. We explore this through "pause tokens," a mechanism that allows for controlled modulation of computational budget during generation, effectively providing the model with more "thinking time" for intricate tasks.

\begin{figure}[htbp] 
    \subfloat[\centering Inference Time Scaling]{%
        \includegraphics[, width=0.8\textwidth, keepaspectratio]{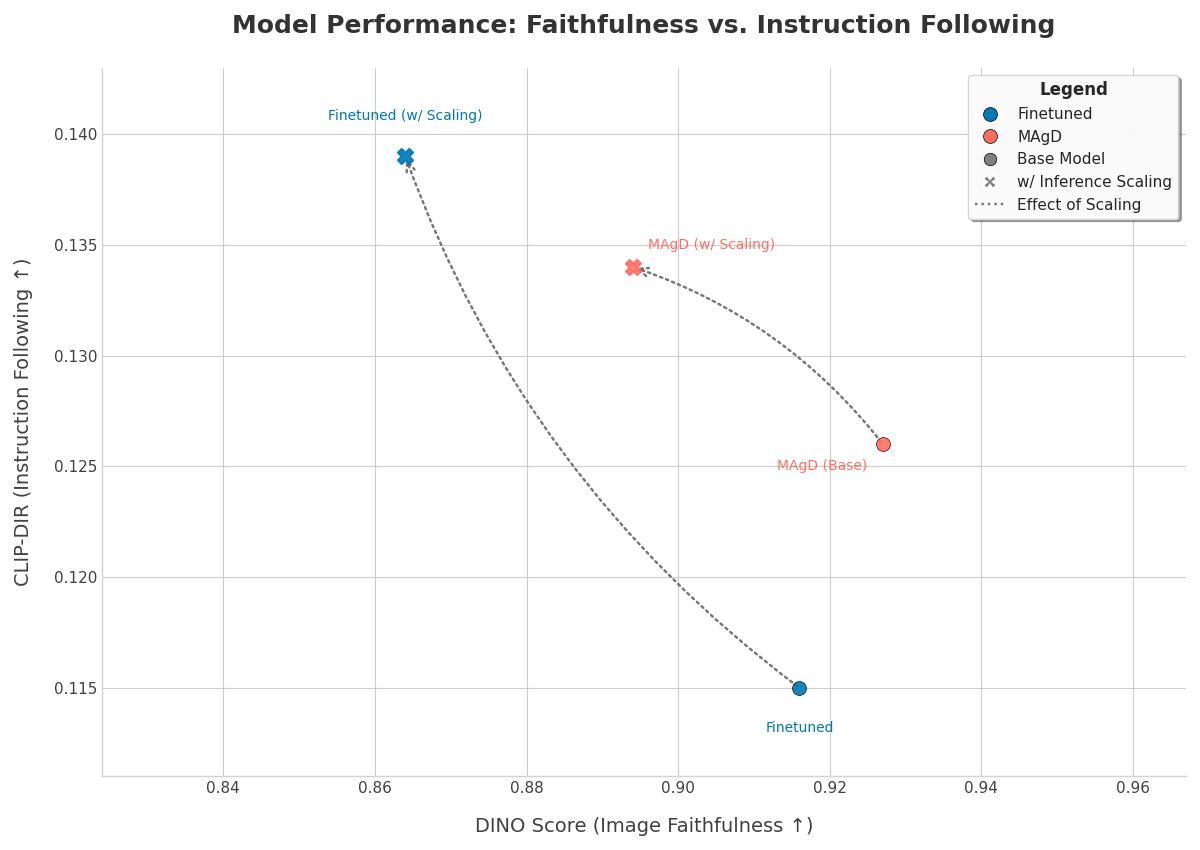}%
    }%
    \caption{Compared to vanilla inference, choosing best CLIP-DIR among different generations with varying pause tokens = 0,8,16,32 we can see significant boost in CLIP-DIR with some loss in faithfulness (DINO) Score}
    \label{fig:scaling_boost_total}%
\end{figure}

Our findings, summarized in Fig.~\ref{fig:scaling_boost_total}, reveal that optimizing for instruction following (measured by CLIP-DIR, which we find better correlates with human perception than CLIP-T) by varying the number of pause tokens (at a fixed seed) yields substantial gains. Specifically, for a checkpoint finetuned on expressive prompts:a model potentially better equipped to leverage additional capacity, \textbf{CLIP-DIR scores surge from $\mathbf{0.115 \to 0.139}$}. This suggests enhanced grounding and exploitation of the augmented capacity. Similarly, the MAgD objective sees its CLIP-DIR improve from $0.126 \to 0.134$, with only a marginal decrease in faithfulness (DINO dropping from 0.93 to 0.89). These results strongly support our initial hypothesis: providing additional inference-time capacity directly translates to better instruction adherence.

\begin{figure}[htbp]
    \centering
    \includegraphics[width=1.0\linewidth]{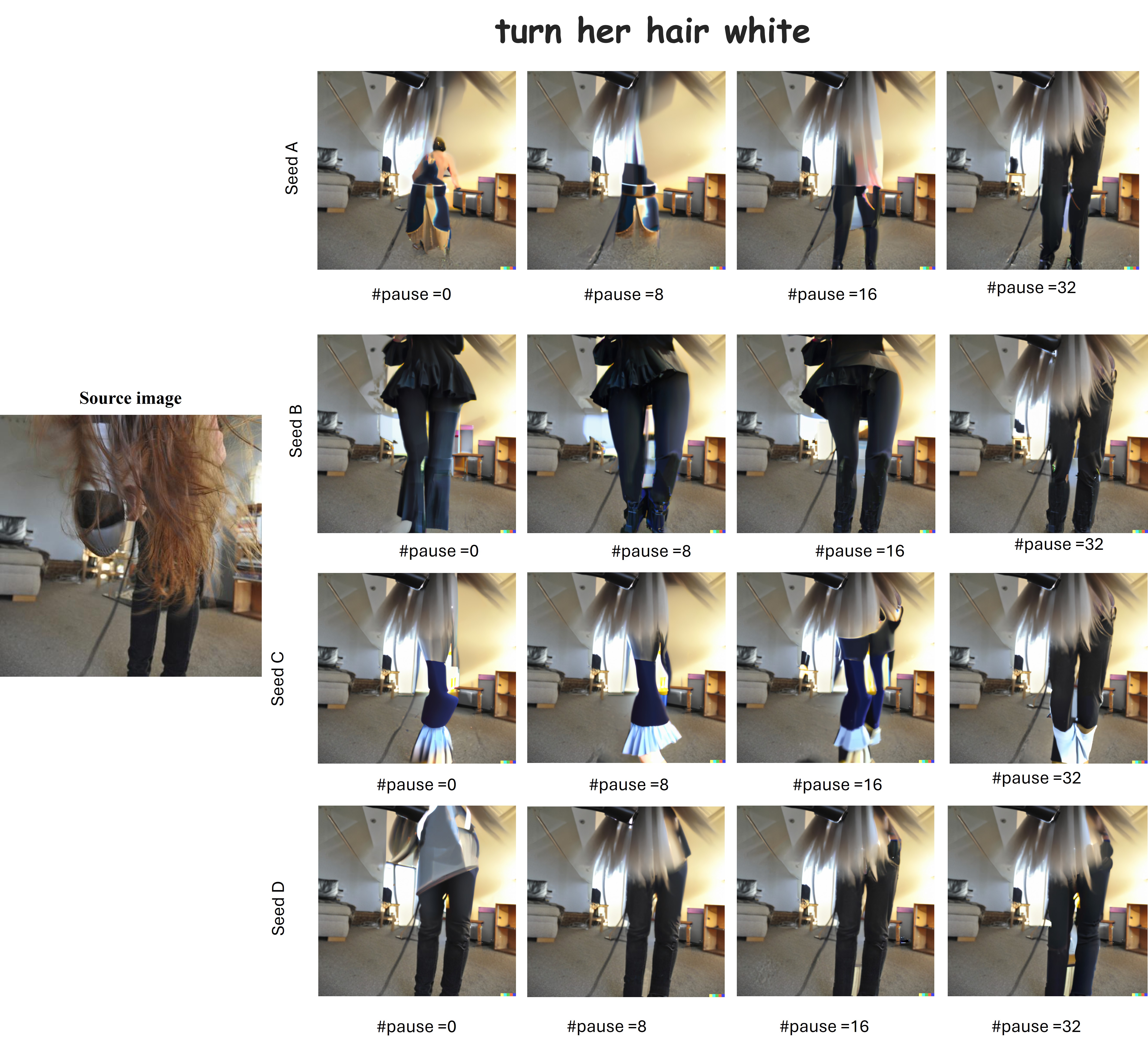}
    \caption{
    \textbf{Effect of think tokens vs different seeds} The figure illustrates the effect of varying the number of pause tokens on an image editing task, prompt: `turn her hair white.' It presents a `Source image' , edited images for different seeds showing the results of the edit with different numbers of pause tokens ($\#pause = 0, 8, 16, 32$).A key observation is that increasing the number of think tokens generally leads to improved compositional fidelity, allowing the model to better integrate the edit with the original scene's structure. This is particularly evident in the pants region: as the number of think tokens increases across each row, the rendering of the pants more closely resembles their appearance in the source image. 
    }
    \label{fig:think_bs_seed}
\end{figure}

\paragraph{Qualitative Analysis:} Fig.~\ref{fig:think_bs_seed} shows that as the number of pause tokens increases, the model's ability to copy the source image while adhering to the target instruction improves. This validates our intuition that additional capacity in visual editing \textbf{facilitates the selective copying and updating of features and attributes}, leading to improved grounded text-based visual editing.

\vspace{-5pt}
\captionsetup{font=small}

\paragraph{Strategic Trade-offs: Maximizing Instruction Following under Faithfulness Constraints.}
Real-world applications often require a delicate balance between accurately following an edit instruction and preserving the original image's salient features (faithfulness). Our framework readily accommodates such needs. 

We propose a practical, inference-time protocol: Generate multiple candidate images by varying the number of pause tokens, filter these candidates to retain only those satisfying a predefined minimum faithfulness threshold (e.g., a DINO score target).From this filtered set, select the candidate image exhibiting the highest instruction-following metric (e.g., CLIP-DIR).

\begin{figure}[htbp] 
    \centering
    \subfloat[\centering ADD Task]{%
        \includegraphics[, width=0.48\textwidth, keepaspectratio]{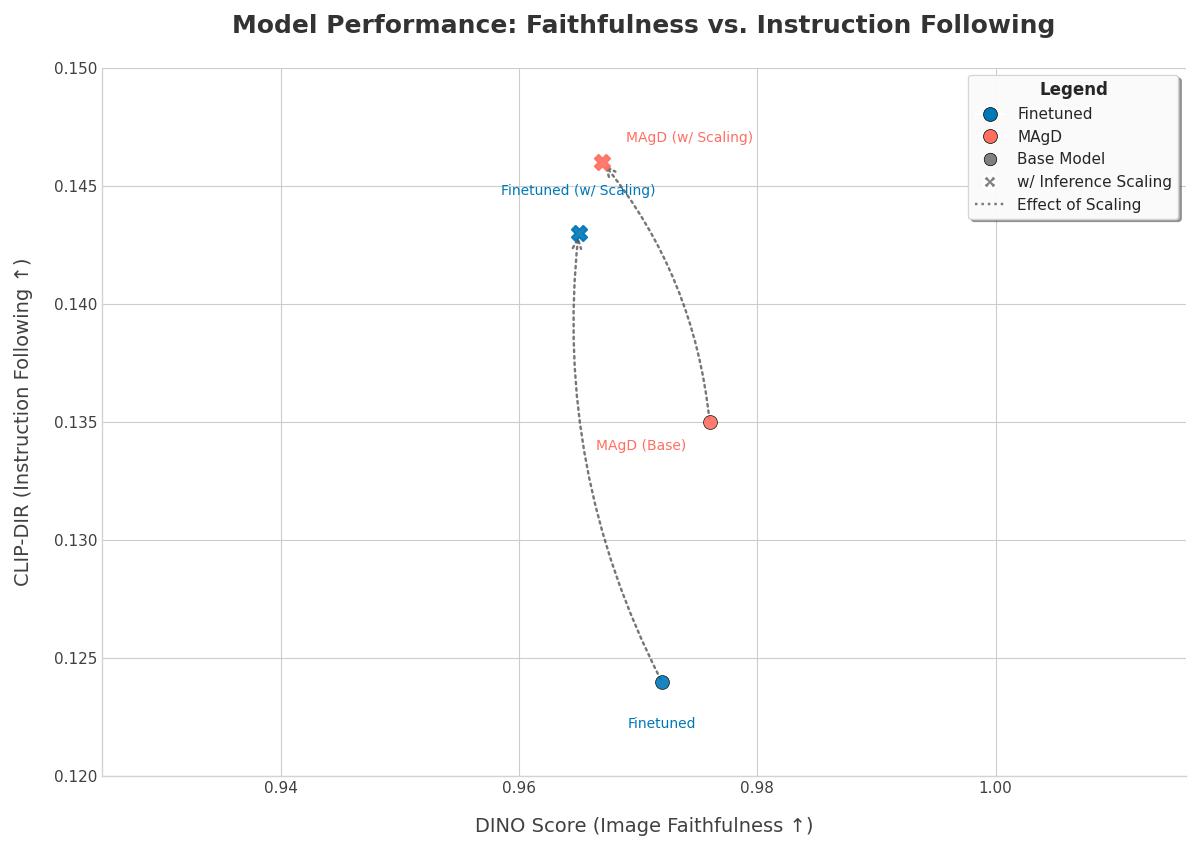}%
    }%
    \hfill 
    \subfloat[\centering Remove Task ]{%
        \includegraphics[, width=0.48\textwidth, keepaspectratio]{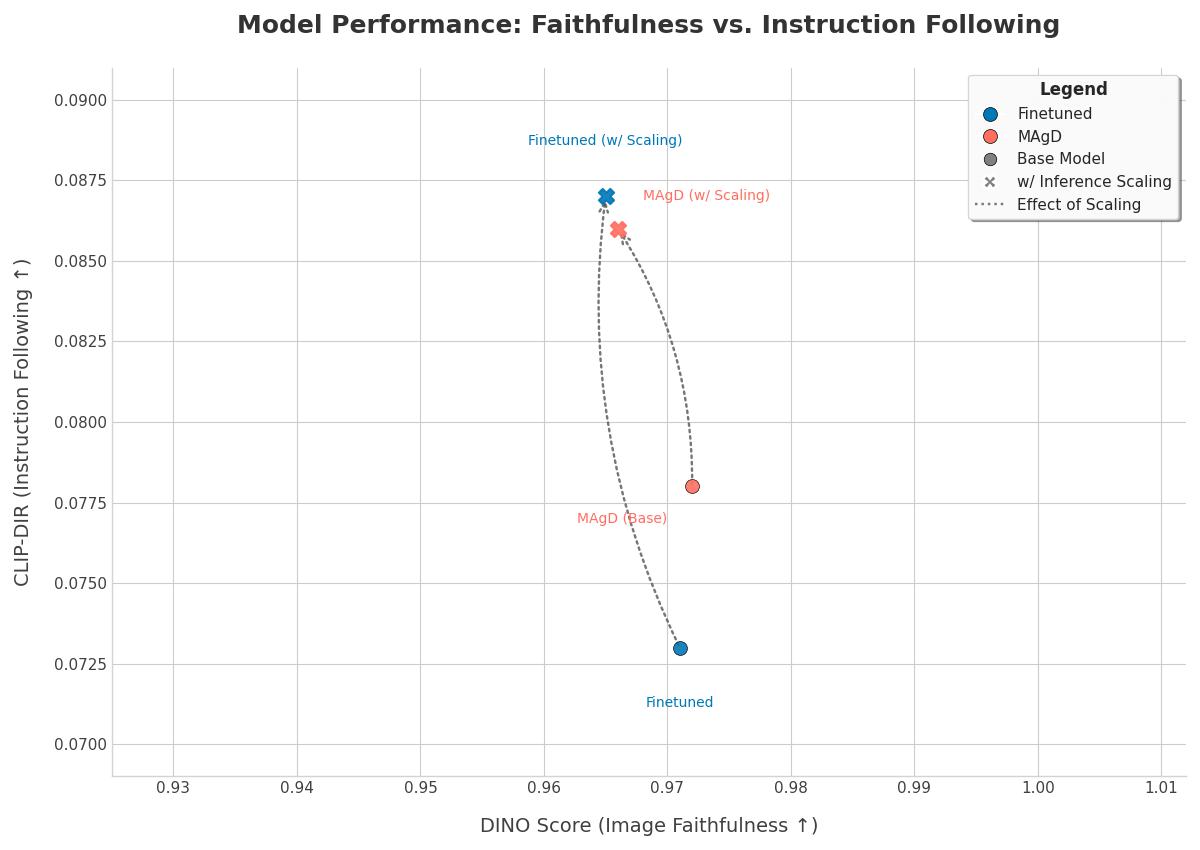}%
    }%
    \caption{Compared to vanilla inference, choosing best CLIP-DIR among different generations with varying pause tokens = 0,8,16,32 we can see significant boost in CLIP-DIR with constraint on faithfulness i.e., (DINO) Score to be $\geq 0.91$}
    \label{fig:add_remove_scaling}%
\end{figure}

\begin{figure}[htbp] 
    \centering
    \subfloat[\centering Global Task]{%
        \includegraphics[, width=0.48\textwidth, keepaspectratio]{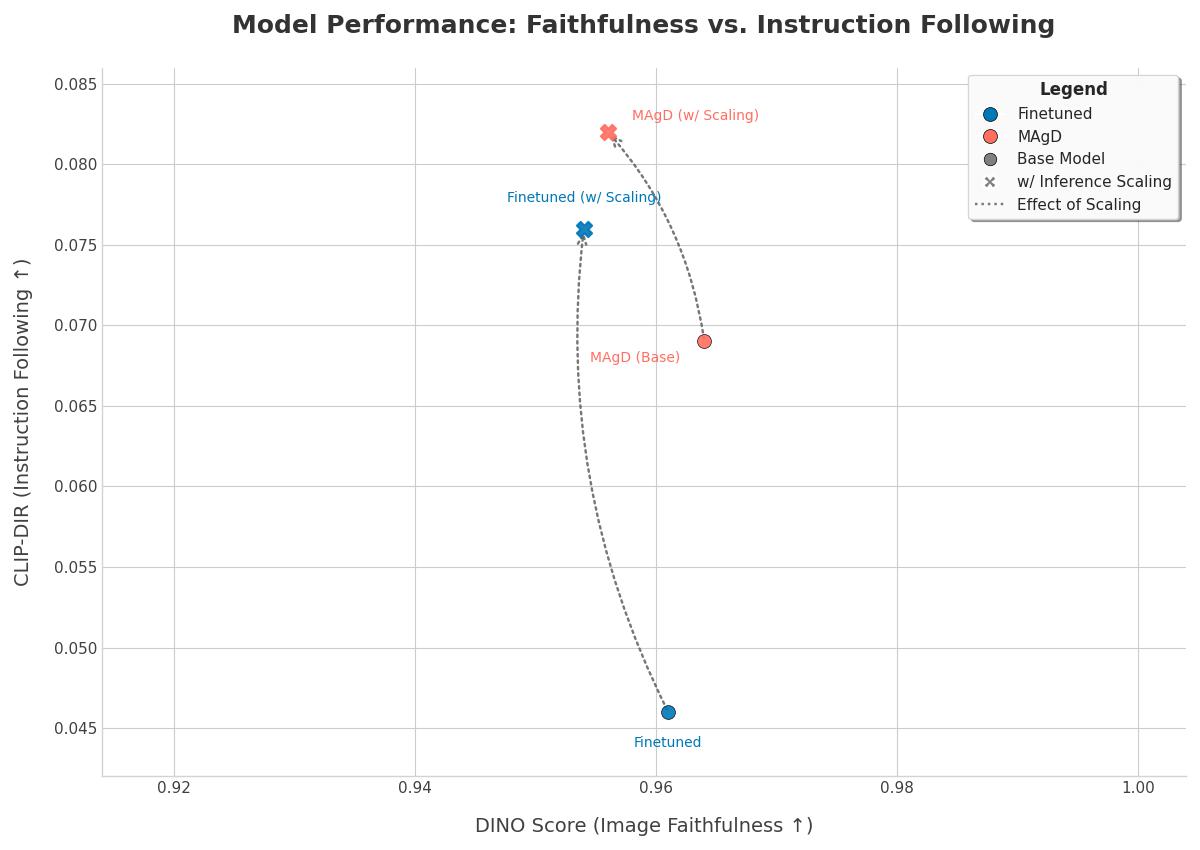}%
    }%
    \hfill 
    \subfloat[\centering Local Task ]{%
        \includegraphics[, width=0.48\textwidth, keepaspectratio]{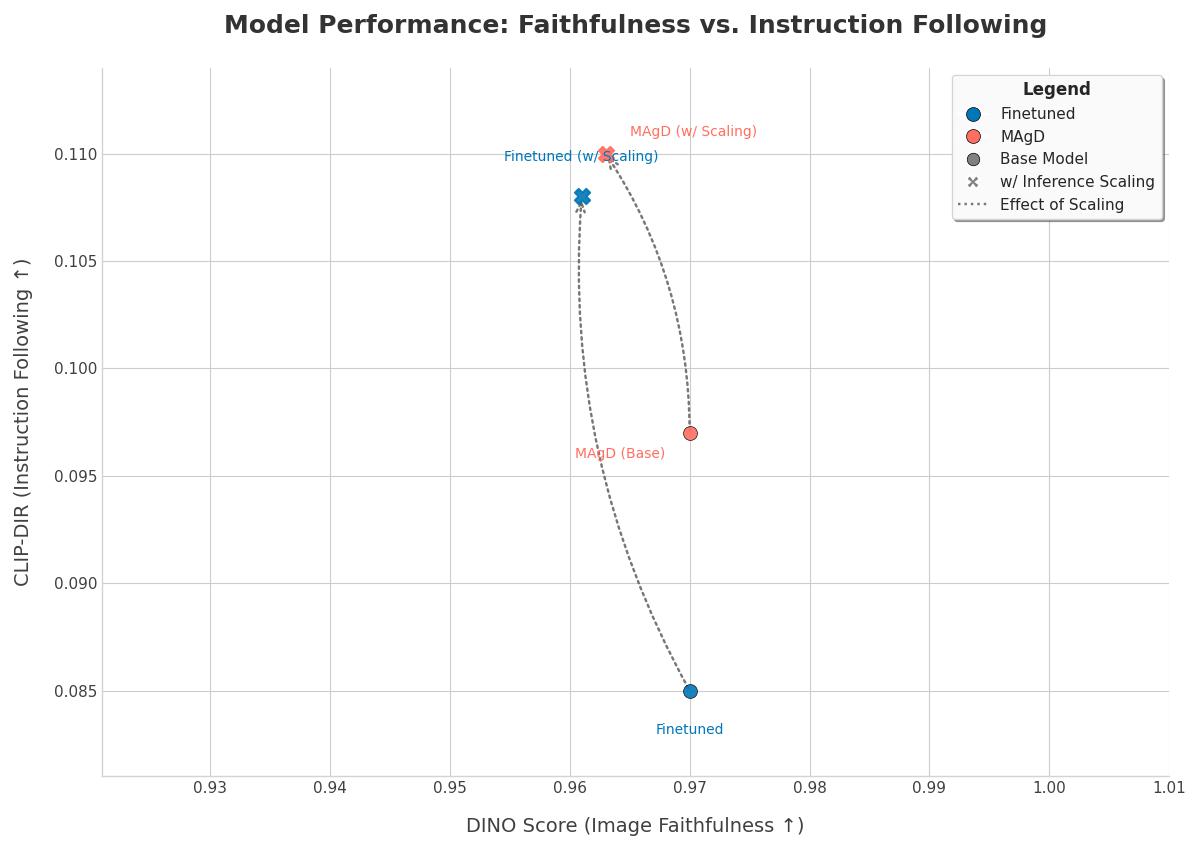}%
    }%
    \caption{Compared to vanilla inference, choosing best CLIP-DIR among different generations with varying pause tokens = 0,8,16,32 we can see significant boost in CLIP-DIR with with constraint on faithfulness i.e., (DINO) Score to be $\geq 0.91$}
    \label{fig:global_local_scaling}%
\end{figure}

This strategy empowers users to navigate the faithfulness-instruction following spectrum effectively. As illustrated in Figs.\ref{fig:add_remove_scaling} and \ref{fig:global_local_scaling}, this constrained optimization protocol delivers striking improvements across diverse editing tasks (add, remove, global, local edits). For the finetuned checkpoint, we observe a consistent boost of over 30\% in instruction following across complexities on Complex Edit Benchmark Tab.\ref{tab:performance_complexity}. The MAgD checkpoint also benefits, achieving approximately a 10\% improvement under similar constraints, indicating that models with appropriate pre-training can effectively utilize this inference-time scaling.

\vspace{-5pt}
\captionsetup{font=small}

\vspace{-5pt}
\captionsetup{font=small}

\begin{table}[htbp]
\centering
\caption{\textbf{Inference scaling by increasing instruction Complexity}. We report best $MLLM_{avg}$ for different complexities with and without scaling. For baseline we report finetuned with Expressive prompt (DP). }
\label{tab:performance_complexity}
\begin{tabular}{@{}>{\raggedright\arraybackslash}p{2.5cm}|llll@{}} 
\toprule
\rowcolor{rowgray}\textcolor{black}{} & \textcolor{black}{\textbf{Complexity=0}} & \textcolor{black}{\textbf{Complexity=1}} & \textcolor{black}{\textbf{Complexity=2}} & \textcolor{black}{\textbf{Complexity=3}} \\
\midrule
Finetuned w/ (DP)& 6.97 & 5.6 & 4.4 & 4.0 \\
\hspace{0.5cm} w/scaling & 8.69 & 7.35 & 6.36 & 5.2 \\
\rowcolor{rowgray} \hspace{0.5cm} $\Delta$ (\%) Improvement & \textit{24.7\%} & \textit{31.3\%} & \textit{44.6\%} & \textit{30.0\%} \\
\midrule
MAgD & 8.166 & 7.269 & 5.5 & 6.25 \\
\hspace{0.5cm} w/scaling & 9.05 & 7.87 & 6.23 & 6.5 \\
\rowcolor{rowgray} \hspace{0.5cm} $\Delta$ (\%) Improvement & \textit{10.8\%} & \textit{8.3\%} & \textit{13.3\%} & \textit{4.0\%} \\
\bottomrule
\end{tabular}
\end{table}

\paragraph{Comparative Analysis on Training objectives.}
An intriguing outcome of inference-time capacity scaling is its impact on the relative performance of different training time objectives. While the finetuned checkpoint significantly benefits, becoming competitive with, or even surpassing, MAgD on 'add,' 'remove,' and 'local edit' tasks, MAgD maintains superior performance for 'global edit' tasks, even when both models leverage pause tokens. This suggests that while inference-time scaling can compensate for some capacity limitations, training objectives like MAgD, designed for improved grounding and compositional visual representation, still hold an advantage for tasks demanding holistic scene understanding and manipulation. Nevertheless, the substantial boost achieved by the finetuned model through mere inference-time scaling underscores the potential for inference time scaling for visual editing.

\paragraph{Inference Scaling on Complex Edit with $\mathbf{MLLM_{avg}}$}
To ensure robust performance evaluation, in Tab.~\ref{tab:performance_complexity} we assign lower weight to samples exhibiting discordant Instruction Following (IF) and Image-Driven Preservation (IDP) scores. Specifically, examples with high IF (IF > 7) and low IDP (IDP < 3) are excluded, as these often reflect hallucination that would bias average results. Similarly, samples with high IDP but low IF (indicating failed instruction execution despite content preservation) are also weighted lower. For the remaining samples, a composite score is computed as (IF+IDP)/2, and the table reports the mean of these scores across all samples

\paragraph{Future Directions for Inference Capacity Scaling:}
In summary, this investigation substantiates the "capacity deficit" hypothesis for complex visual editing and inverse problems within standard diffusion pipelines. We demonstrate that in-context, inference-time capacity scaling, achieved through the simple yet effective mechanism of pause tokens, is a potent and flexible technique to significantly enhance performance. This approach not only improves raw instruction following but also allows for nuanced control over the faithfulness-instruction following trade-off. The promising results open avenues for future exploration, particularly in contexts requiring iterative refinement, multi-turn editing, or handling even more complex compositional instructions, by providing more intermediate steps and corresponding feedback at training time and also potentially during generation.
\\

\section{Ablation Studies and Design Choices in MAgD Training}

To better understand the impact of various design decisions within our proposed Masking-Augmented Gaussian Diffusion (MAgD) framework, we conduct a series of controlled ablation studies. These ablations evaluate the effect of key hyper-parameters and architectural choices on the model’s ability to learn robust, compositional, and semantically grounded representations for visual editing.

In MAgD training, the masking-based auxiliary corruption is applied stochastically with a probability $p_{magd}$ at each optimization step, following a classifier-free guidance formulation. A uniform random variable $u \sim \mathcal{U}(0,1)$ is sampled to determine whether the dual corruption (masking + noise) is applied for a given training instance. Crucially, since our objective is to enhance contextual and compositional representations, we restrict the application of the masking operation to higher noise levels i.e., when the denoising network primarily focuses on low-frequency structural components. This behavior is governed by a time-step threshold parameter $\tau_{MAgD} \in [0,1]$, such that masking is applied only when the diffusion time-step $ t \geq \tau_{MAgD}$ (high noise level). $r_{mask}$ is our fixed maskingn rate we use to obtain modified masked noisy input $\tilde{x}^{masked}_t$. Our overall objective is:
\begin{equation}
\label{eqn:objective}
\mathcal{L}_{\text{MAgD}} = 
\begin{cases}
   \mathcal{L}_{\text{mDSM}}(\tilde{\mathbf{x}}_t^{\text{masked}}, \epsilon_\theta), & \text{if } u < p_{\text{magd}} \text{ and } t < \tau_{\text{MAgD}} \\
    \mathcal{L}_{\text{DSM}}(x_t, \epsilon_\theta), & \text{otherwise}
\end{cases}
\end{equation}


\begin{table}[htbp]
\centering
\caption{\textbf{MAgD objective ablations:} Performance metrics as a function of varying the masking application strategy. We ablate (1) the proportion of training steps where masking is applied (for timesteps $t < \tau_{MAgD}$) and (2) the mask rate $r_{mask}$ (percentage of tokens dropped).}
\label{tab:magd_params_ablation}
\resizebox{\textwidth}{!}
{
\begin{tabular}{@{}>{\raggedright\arraybackslash}p{4cm} l l l l l l l@{}} 
\toprule
\rowcolor{headergray} \textbf{}  & \textbf{$r_{mask}$} & $\tau_{\text{MAgD}}$ & \textbf{Clip-I} & \textbf{DINO} & \textbf{CLIP-DIR} & \textbf{CLIP-T} & $MLLM_{avg}$ \\ 
\midrule
MAgD & 0.25 & $\geq 0.7$ & 0.854 & 0.899 & 0.1218 & 0.2609 & 8.3 \\
\hspace{0.5cm} Higher \% $\tau_{\text{MAgD}}$  & 0.25 & $\geq 0.5$ & 0.822 & 0.853 & 0.1315 & 0.258 & 7.98\\
\hspace{0.5cm} Higher $r_{mask}$ & 0.5 & $\geq 0.7$ & 0.859 & 0.898 & 0.1233 & 0.259 & 8.1\\
\bottomrule
\end{tabular}
}
\end{table}


\paragraph{Noise level Threshold Selection}
In our formulation (Eq.~\ref{eqn:objective}), $\tau_{\text{MAgD}}$ determines the noise-level threshold above which masking-based secondary corruption is applied. Our core intuition is that masking should target high noise levels where the denoising network focuses more on reconstructing semantic and low-frequency components, which are crucial for capturing global structure and compositionality.

During training, masking-based corruption is applied with 50\% probability and only at timesteps $t \geq \tau_{\text{MAgD}}$. Effectively, this results in dual corruption being active for only a fraction $0.5 \times \tau_{\text{MAgD}}$ of training steps. In Tab.~\ref{tab:magd_params_ablation}, we compare two settings of $\tau_{\text{MAgD}}$ corresponding to dual corruption being applied over roughly 30\% vs. 50\% of the diffusion forward process.

Our results show that both settings lead to competitive performance when paired with a moderate masking rate, but a lower threshold ($\tau$) yields more stable training, acting more like a form of semantic regularization. .

\paragraph{Masking rate $\mathbf{r_{mask}}$ Ablation}
We explore two settings for the masking rate: 25\% and 50\% to assess the trade-off between supervision strength and stability. As reported in Tab.~\ref{tab:magd_params_ablation}, both configurations produce comparable results in our EmuEdit evaluation.

However, we observe that a lower masking rate (25\%) leads to slightly more consistent improvements across tasks and training runs. Given our intent to use the dual corruption as a regularization mechanism, a lower masking rate proves to be both effective and stable.


\paragraph{Efficacy of Learnable Mask Embeddings}
We also investigate both with learnable mask embedding or zero-value for masked tokens when dual corruption is adopted within MAgD training.

Learnable mask follows exact design as time-step embedding in Omnigen and consists of two layer MLP. We initialize a pool of  1024 learnable mask tokens with a hidden dimension of 256. For each timestep we stochastically sample from one of the mask embeddings. For fair comparison the composite objective kicks in exact setting at high noise-levels i.e., $t \geq 0.7$  at a  masking rate $r_{mask} = 0.25$. 

From Tab.~\ref{tab:Learnt mask ablation} ablation on EmuEdit evaluates both with and without  learnable mask within MAgD training setup and both checkpoints are trained with a cosine schedule after a warmup stage and generations are done for ~3200 gradient steps. 

\begin{table}[htbp]
    \centering
    \begin{tabular}{lcccc}
        \toprule
        \rowcolor{headergray} & CLIP-I & CLIP-T & DINO & CLIP-DIR \\
        \midrule
        Zero masking  & 0.855 & 0.265 & 0.915 & 0.123  \\
        Learnable Mask Embedding  & 0.849 & 0.268 & 0.913 & 0.127 \\
        \bottomrule
    \end{tabular}
    \caption{\textbf{Learnable Mask Embedding Ablation} We can observe that while both work, we can observe better CLIP-DIR with learnable mask embedding.}
    \label{tab:Learnt mask ablation}
\end{table}

\section{Limitations of Metrics and Qualitative results}
Our discussion on the importance of robust evaluation metrics is presented in two sections. Initially, we introduce the chosen metrics and elaborate on their respective relevance for each task. Subsequently, we detail the inherent limitations of metrics

\paragraph{Image-Editing Metrics}
For our evaluations we report two metrics for faithfulness CLIP-I, DINO scores. Although CLIP understands the semantics better, it does not account for the actual appearance or unrealistic nature of the edit. However while DINO provides better pixel level fidelity does not have capture the global representations well. The efficacy of these metrics are also dependent on the tasks at hand. For Text to image alignment we report CLIP-T and directional CLIP similarity.It is known in litterature that for editing CLIP-T may not be sufficient and robust. We report Directional clip similarity CLIP DIR as a more robust measure to track both the scene compositionality given the source image and text-image alignment given the source and target prompt. 

\begin{figure}[htbp]
    \centering
    \includegraphics[width=1.0\linewidth]{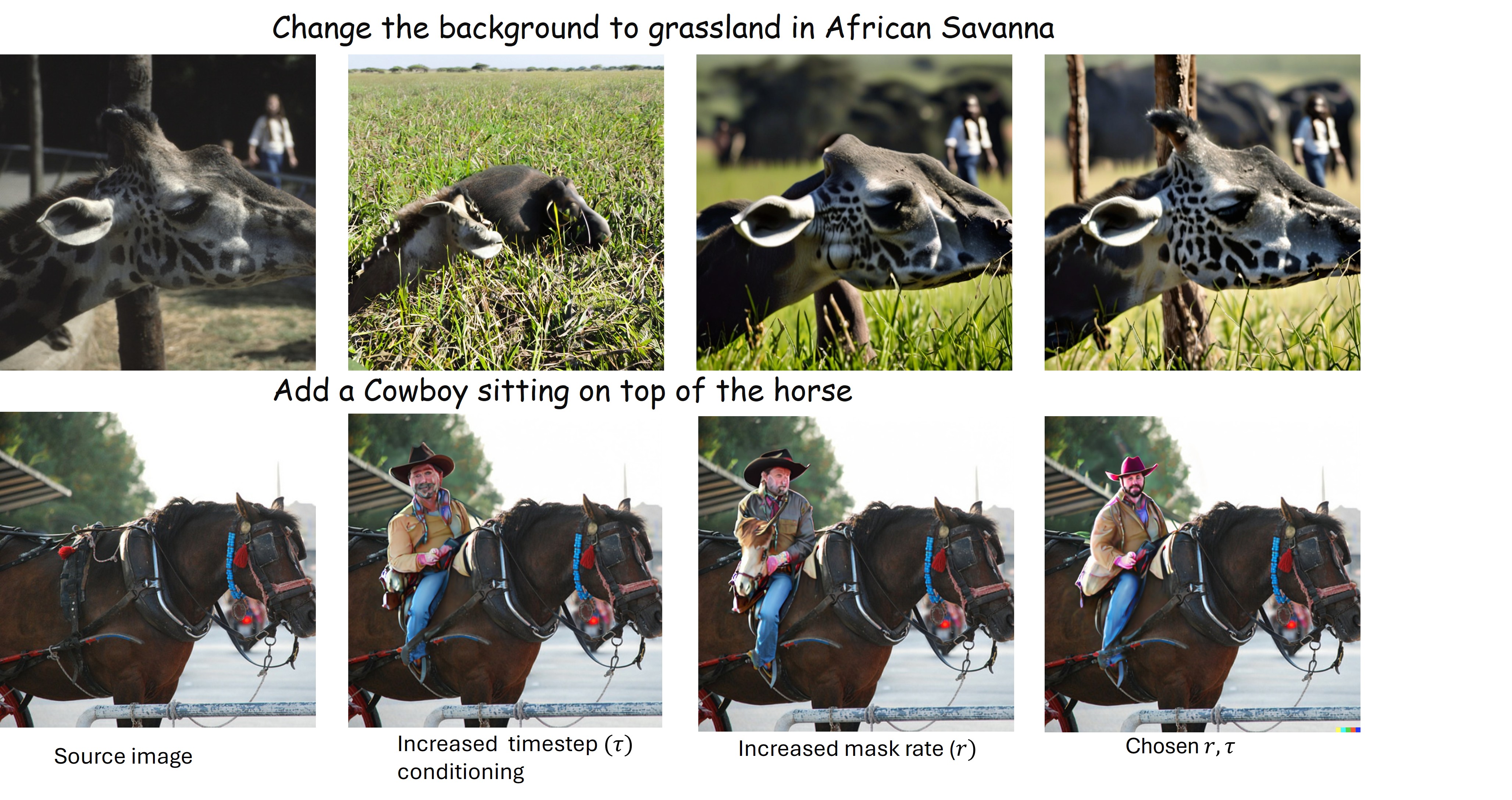}
    \caption{\textbf{Qualitative visualizations across different masking rate and noise level threshold}.This figure provides qualitative examples illustrating the influence of the noise level threshold $\tau$ and masking rate (r) on image editing quality. The top row (Giraffe example) for "Change the background to grassland in African Savanna" shows that increasing the noise level threshold through longer timestep conditioning ($\tau < T - 0.5$) can improve instruction following but may introduce more global distortions in the scene composition.The bottom row (Horses example) for "Add a Cowboy sitting on top of the horse" compares outputs under different conditions: Specifically, increasing the mask rate (e.g., from 0.3 to 0.5) reduces the number of unmasked priors, which can lead to localized distortions (e.g., misplaced horse head). These visualizations highlight the trade-offs between instruction adherence and visual fidelity under varying masking strategies.}
    \label{fig:enter-label}
\end{figure}

\begin{figure}[htbp]
    \centering
    \includegraphics[width=\linewidth]{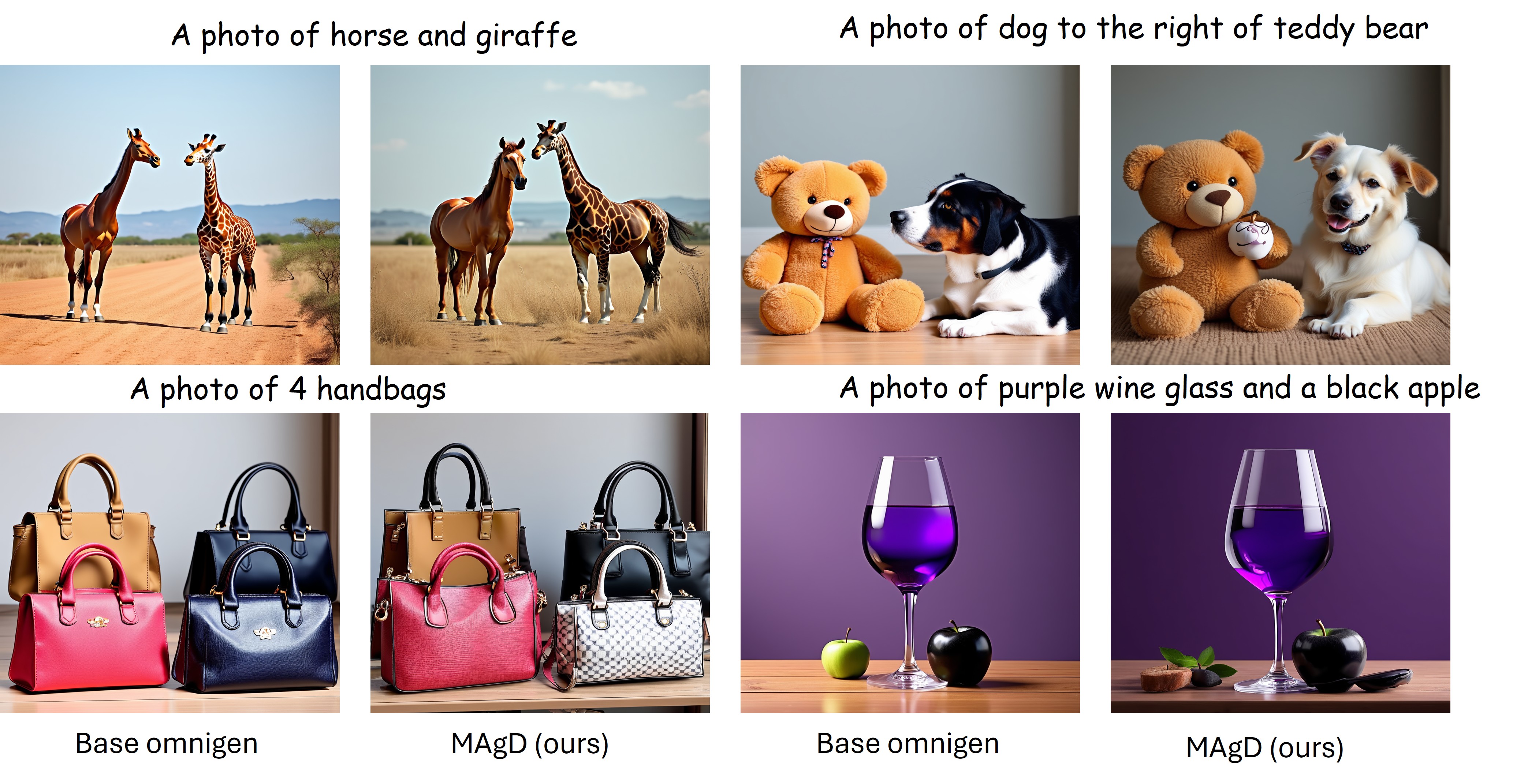}
    \caption{\textbf{Our method on Text-Image generation tasks}. We visualize the efficacy on our model on T2I tasks. Across two object, counting, colors and position the Base Omnigens performance does not deteriorate.}
    \label{fig:enter-label}
\end{figure}

\begin{figure}[htbp]
    \centering
    \includegraphics[width=0.9\linewidth]{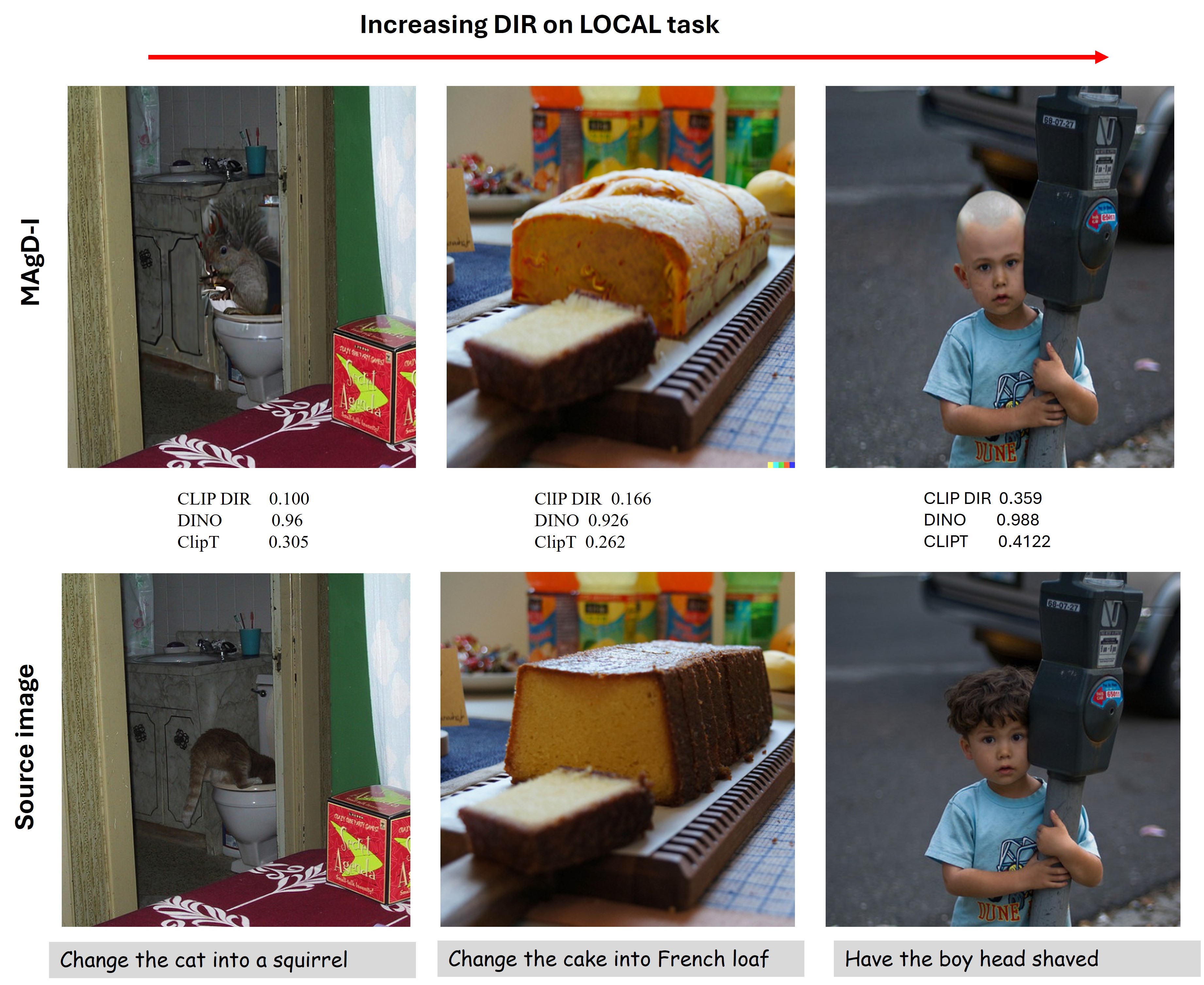}
    \caption{\textbf{This figure presents image generations with progressively increasing CLIP DIR for LOCAL task} (a) Squirrel: While visually accurate and semantically aligned, this generation exhibits a low CLIP DIR despite a high CLIP-T score. (b) \& (c) Cake-to-Bread Transformation \& shaved head transformation: These examples illustrate a semantic transformation task. CLIP DIR is lower since a small "piece of cake" in (b) remains distinct and has not fully transformed into "bread,".}
    \label{fig:increasing dir}
\end{figure}

\paragraph{CLIP DIR vs DINO on different tasks}
We depict the visualizations for interpreting the CLIP-DIR and DINO scores. We motivate that these scores have different statistics and are subjective to the task. For example, in \ref{fig:increasing dir}, depicted for LOCAL task of emuedit which performs region based edits, the DINO scores are typically very high compared to background editing or GLOBAL edits etc. Hence the average dino score for global editing task might be different from region based editing task or Image enhancement tasks such as stylization, relighting etc. MLLM scores, appear more robust to local or global edits compared to CLIP metrics.
Below, are visualizations for increasing directional similarity for background tasks.
\begin{figure}[htbp]
    \centering
    \includegraphics[width=0.9\linewidth]{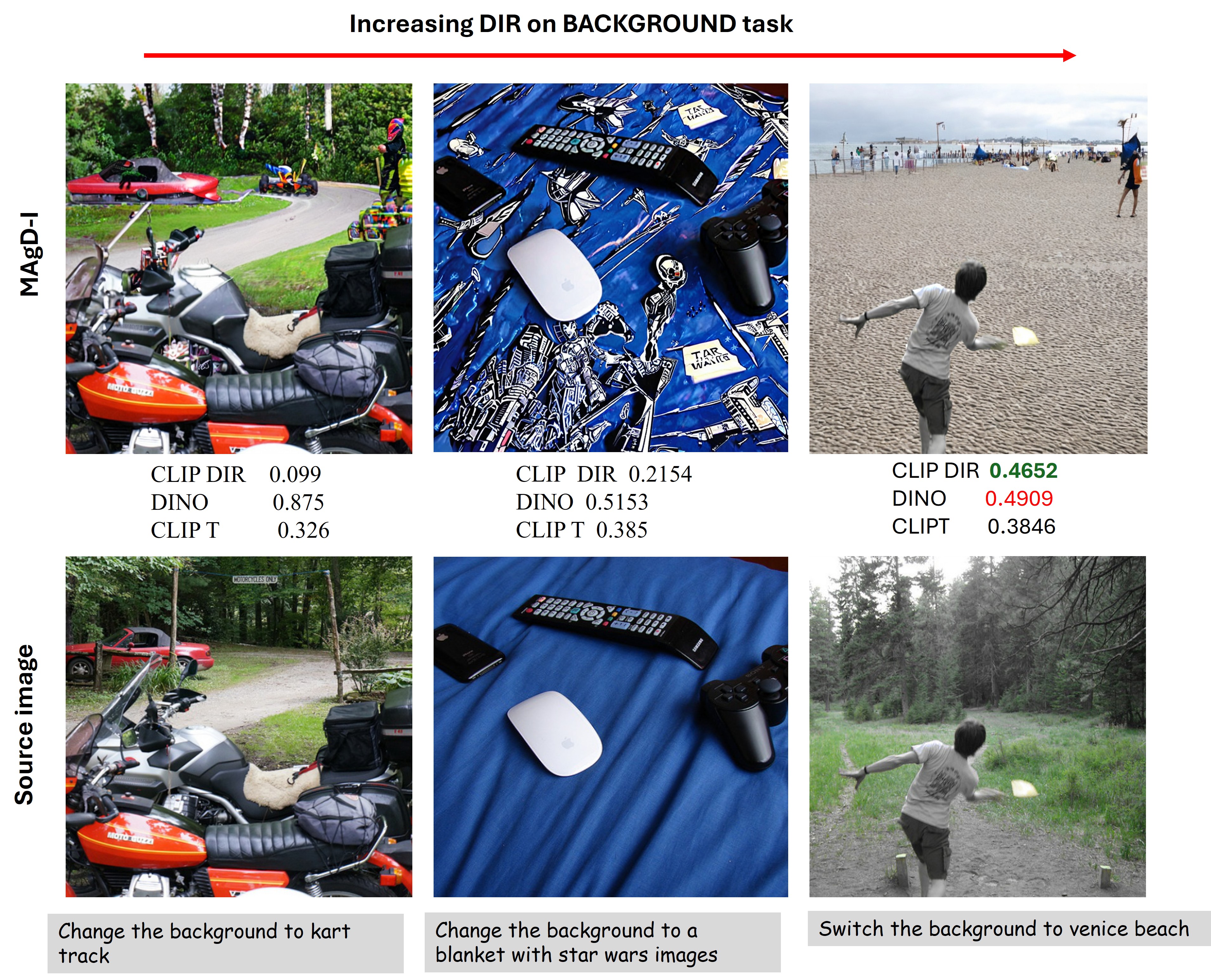}
    \caption{\textbf{This figure presents image generations with progressively increasing CLIP DIR for Background task}. CLIP DIR accurately reflects the scores based on generations. Its important to notice that DINO scores for these tasks are generally low, a higher DINO scores might mostly be accompanied with lower CLIP DIR reflecting lack of instruction following.DINO scores depend on the foreground background composition of each image. Hence}
    \label{fig:increasing dir 2}
\end{figure}

\begin{figure}[htbp]
    \centering
    \includegraphics[width=0.7\linewidth]{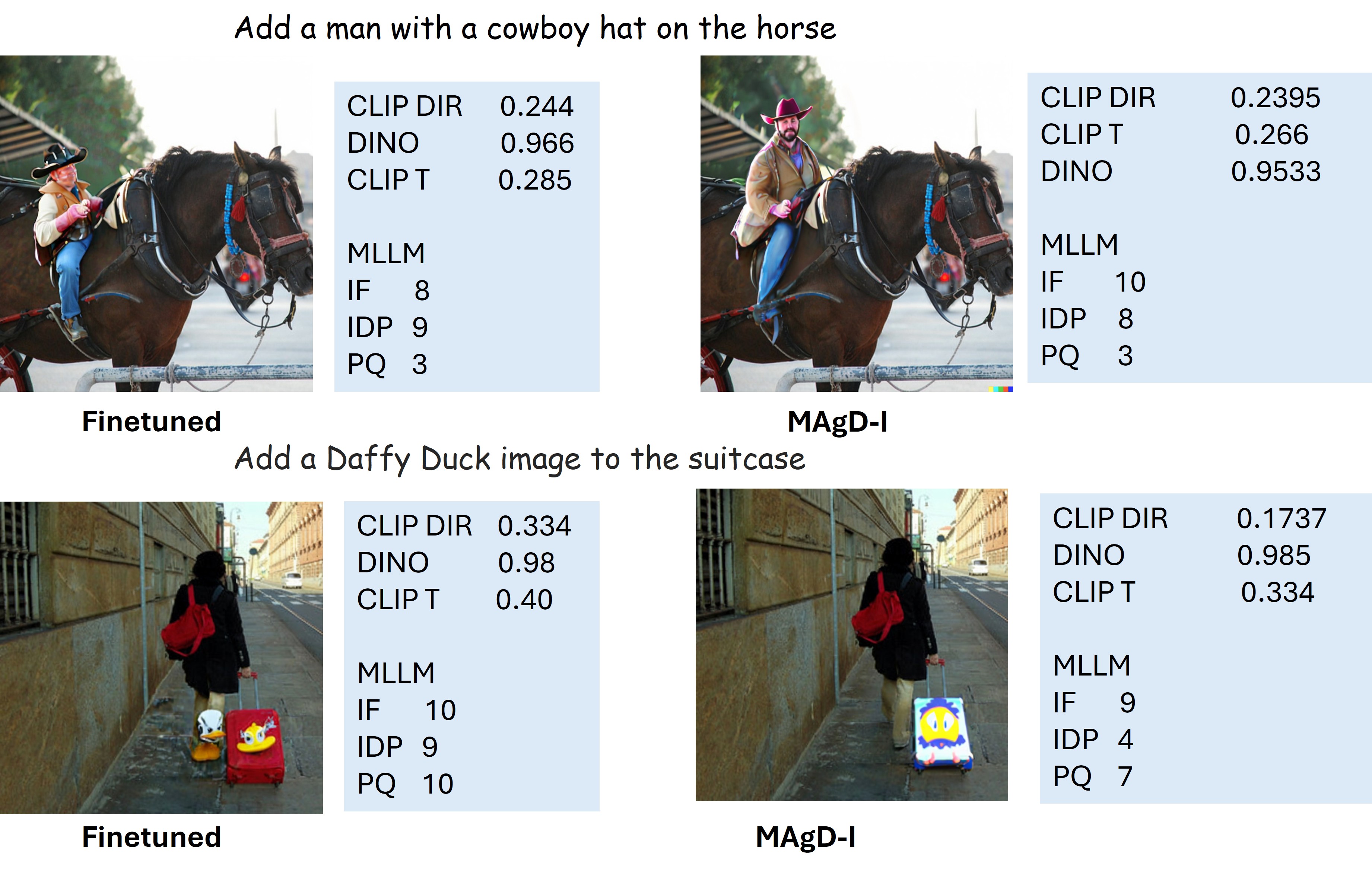}
    \caption{\textbf{Metrics comparisons between Finetuned generations and MAgD-I}. Top: Although both generations add a cowboy on the hors, MAgD-I is perceptually better composed into the image. However CLIP DIR scores are lower compared to the finetuned image. CLIP-T scores also being lower. [Bottom] While finetuned follows the instruction to add the duck into the image there are additional artifacts generated (another face of a duck is generated) however, metrics don't reflect that.}
    \label{fig:CLIP variance}
\end{figure}

\begin{figure}[htbp]
    \centering
    \includegraphics[width=0.8\linewidth]{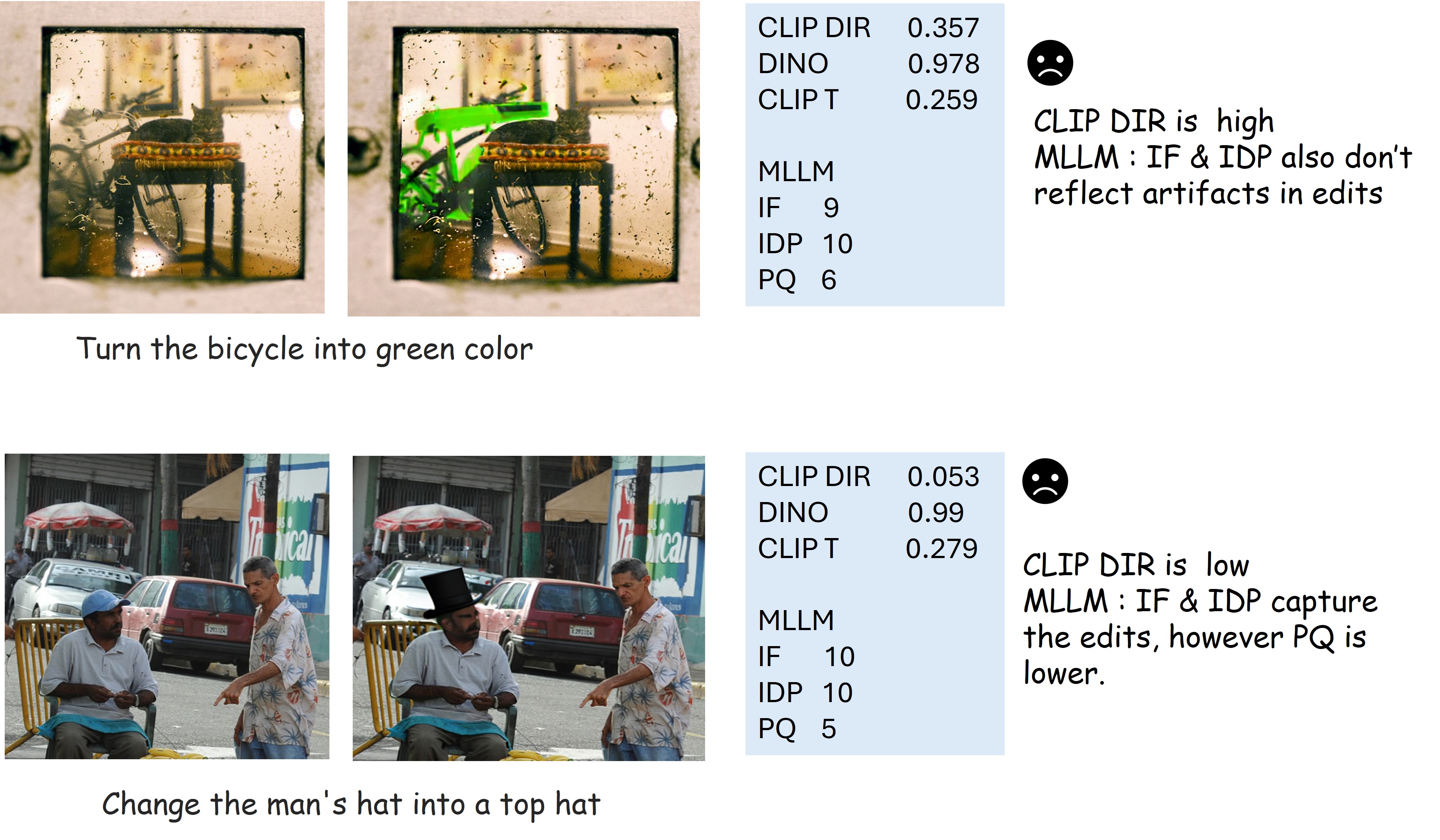}
    \caption{\textbf{Examples of CLIP DIR not reflecting the edits}: Top: For the color task the image generated adds artifacts while not really reflecting the bike structure. However, the CLIP DIR is more biased towards instruction. Here Instruction following metric and Faithfullness also do not follow.  Bottom: Here the editing is accurate, however CLIP DIR is very low not reflecting the edit. MLLM is however, reflective making it more important observe MLLM and DIR scores for editing.}
    \label{fig:More on clip variance}
\end{figure}

\paragraph{ Limitations of CLIP DIR}
CLIP-Dir despite providing a balanced metric that is robust to hallucinations still is measured in CLIP semantic space making it less robust to edited changes where semantic relationships are not preserved, perceptual quality etc. We highlight the issues in two figures below \ref{fig:CLIP variance} and \ref{fig:More on clip variance}. In future work, this warrants for a better model that can capture hallucinations, better object relationships and composition. We observe that in \ref{fig:More on clip variance}, CLIP DIR is more biased towards the instruction not considering the pixel level details, where the structure of the edit does not follow the structure of the bike in the figure. To alleviate these problems we also evaluate using MLLMs specifically we use Gemini Flash 2.0 for evaluations. 


\paragraph{MLLM as judge when CLIP DIR fails}
In conjuction with CLIP DIR we use Gemini Flash to generate Questions $\&$ Answers with corresponding ratings for every Source image and prompt instruction pair.This is shown in the figure \ref{fig:mllm_eval_box}. We prompt gemini to generate five Questions that evaluate the generations on 1. " \emph{if the generated image aligns with edit prompt ?}", 2. "\emph{if the edited image is faithfull to the prompt and source image?}"  and 3. \emph{"if the edited image has no distortions, artifacts, lightning and shadows are blending with the background.?}". Gemini then generate questions and answers for each prompt, image pair. For each model, we then probe an MLLM to choose one answer which corresponds to a reasoning for each question given a (generated image, instruction) pair. The questions are tagged as 'IF' ( Instruction following), 'PQ' (Perceptual quality and 'IDP' (  Identity preservation) similar to complex-edit \cite{yang2025textttcomplexeditcotlikeinstructiongeneration}. We further weight these answers on scale of 1-10. We observe that provides better robustness to relationships and object quality and spatial attribute compared to CLIP-DIR. Hence we use MLLM scores in conjuction with CLIP DIR. For reference in \ref{fig:More on clip variance}, In attribute editing, IF and IDP scores align well with qualititative observations. In color task, the Perceptual quality is lower and IF as well as the bike is not colored accurately. We admit, however the quality of MLLM scores and robustness an be further improved and that is left for future work.

\begin{figure}[htbp]
\centering
\begin{tcolorbox}[title=MLLM Evaluation Benchmarking Example, colback=gray!5, colframe=gray!40!black, fonttitle=\bfseries, width=\textwidth]
\textbf{Question 1:} How well does the generated image align with the edit prompt \textit{"Make the sky night instead of day"}?\\
\textbf{Answers:}
\begin{itemize}
    \vspace{-4pt}
    \item[0:] No alignment; the image still clearly depicts daytime.
    \vspace{-2pt}
    \item[1:] Minimal alignment; slight darkening, still mostly daytime.
    \vspace{-2pt}
    \item[2:] Partial alignment; suggests twilight, but not fully night.
    \vspace{-2pt}
    \item[3:] Moderate alignment; sky is dark but could be enhanced.
    \vspace{-2pt}
    \item[4:] Good alignment; dark sky, but some daytime elements remain.
    \vspace{-2pt}
    \item[5:] Excellent alignment; sky clearly depicts nighttime.
    \vspace{-2pt}
\end{itemize}

\textbf{Question 2:} How faithfully does the edited image preserve the original content while making the sky night?\\
\textbf{Answers:}
\begin{itemize}
    \vspace{-4pt}
    \item[0:] Completely unrecognizable due to extensive changes.
    \vspace{-2pt}
    \item[1:] Significant portions are altered or missing.
    \vspace{-2pt}
    \item[2:] Some key elements preserved, but substantial changes.
    \vspace{-2pt}
    \item[3:] Most key elements preserved, minor alterations.
    \vspace{-2pt}
    \item[4:] Key elements preserved; sky change doesn’t degrade quality.
    \vspace{-2pt}
    \item[5:] Perfect preservation with only sky changed convincingly.
    \vspace{-2pt}
    \item[-1:] Cannot determine.
    \vspace{-2pt}
\end{itemize}

\textbf{Question 3:} To what extent does the edited image exhibit distortions, artifacts, or lighting inconsistencies?\\
\textbf{Answers:}
\begin{itemize}
    \vspace{-4pt}
    \item[0:] Severe distortions and jarring effects.
    \vspace{-2pt}
    \item[1:] Noticeable artifacts that detract from quality.
    \vspace{-2pt}
    \item[2:] Minor, non-distracting artifacts.
    \vspace{-2pt}
    \item[3:] Minimal distortions visible only on close inspection.
    \vspace{-2pt}
    \item[4:] Very few artifacts; nighttime lighting blends well.
    \vspace{-2pt}
    \item[5:] No noticeable issues; night effect blends seamlessly.
    \vspace{-2pt}
    \item[-1:] Cannot determine.
    \vspace{-2pt}
\end{itemize}

\textbf{Question 4:} Which adjustment best contributes to the illusion of nighttime?\\
\textbf{Answers:}
\begin{itemize}
    \vspace{-4pt}
    \item[0:] Increased brightness and contrast.
    \vspace{-2pt}
    \item[1:] Addition of daytime clouds.
    \vspace{-2pt}
    \item[2:] Twilight-simulating color hues.
    \vspace{-2pt}
    \item[3:] Replacing the sun with the moon.
    \vspace{-2pt}
    \item[4:] Reduced brightness and blue/purple hue adjustment.
    \vspace{-2pt}
    \item[-1:] Cannot determine.
    \vspace{-2pt}
\end{itemize}

\textbf{Question 5:} How realistic and plausible is the lighting and shadow in the edited image?\\
\textbf{Answers:}
\begin{itemize}
    \vspace{-4pt}
    \item[0:] Completely unrealistic for nighttime.
    \vspace{-2pt}
    \item[1:] Largely unrealistic with strange effects.
    \vspace{-2pt}
    \item[2:] Somewhat plausible with inconsistencies.
    \vspace{-2pt}
    \item[3:] Generally plausible with minor issues.
    \vspace{-2pt}
    \item[4:] Highly realistic, convincing nighttime feel.
    \vspace{-2pt}
    \item[5:] Exceptionally realistic, like a real night photo.
    \vspace{-2pt}
    \item[-1:] Cannot determine.
\end{itemize}
\end{tcolorbox}
\vspace{-8pt}
\caption{MLLM-based evaluation questions and answer schema used for assessing alignment, faithfulness, realism, and plausibility in edited generations.}
\label{fig:mllm_eval_box}
\end{figure}

\section{Experimental Setup}
In this section we discuss the hyperparameters for all the experiments and ablations, training dataset and generation of dense prompts. 
\paragraph{Training setup}
For all the experiments we follow same hyperparameters during training as shown in the \ref{tab:hyperparameters}. For Editing tasks we add a weighted loss as followed in omnigen to allow the model to penalize generations where the model simply copies the original image. For T2I task we follow diffusion loss. To train MAgD loss we setup learnable mask as described in earlier sections on Ablations. We allow a increasing noise threshold through the dual corruption process for the first $30\%$  of timesteps. We also apply masked image as a condition with CFG of 0.5. For finetuning and finetuning with expressive prompts the setup remains the same except we don't apply target masking or Dual corruption at the training time. 

\begin{table}[H]
\centering
\caption{Hyperparameters for all for MAgD and Finetuning experiments}
\label{tab:hyperparameters}
\begin{tabular}{@{}>{\bfseries}l@{}>{\columncolor{white}}l@{}} 
\toprule
\rowcolor{headercolor}\textcolor{white}{Hyperparameters} & \textcolor{white}{Value} \\
\midrule
\rowcolor{rowgray}Learning rate & 1e-4 \\
Schedule & Cosine \\
\rowcolor{rowgray}Warmup & 800 \\
Batch size & 128 \\
\rowcolor{rowgray}Gradient steps & $\sim$3200 \\
Resolution & 1024 \\
\rowcolor{rowgray}Training data size & 400k \\
Masking rate & 0.25 \\
\rowcolor{rowgray}Timestep conditioning & T-0.3 \\
Mask embedding size & 1024 \\
\rowcolor{rowgray}Masking probability & 0.5 \\
\bottomrule
\end{tabular}
\end{table}

\begin{figure}[htbp]
    \centering
    \includegraphics[width=1.0\linewidth]{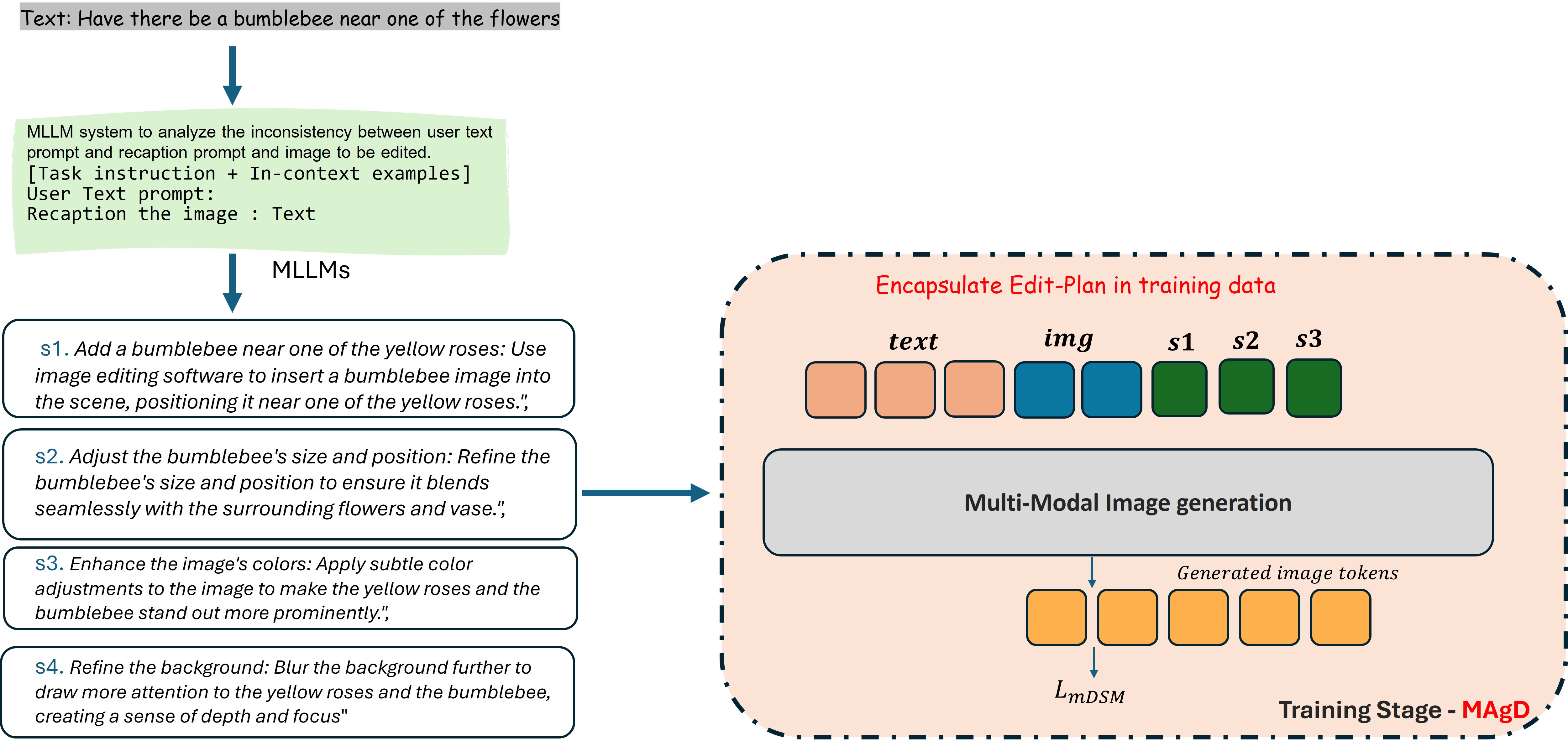}
    \caption{\textbf{Building expressive prompt corpus for Training}: Instruction, source image prompt and source image is passed to the MLLM to analyse the inconsistancy between the text and source prompt and image. Further MLLM is instructed to generate upto 5 sequential concise  prompts splitting the core task into sub-tasks (s1), (s2), (s3), (s4), (s5). Further interleave these prompts in the image embedding, ensuring the image id and embedding is followed by the step instructions generated in the previous step. This is not done during Inference.}
    \label{fig:Expressive generation}
\end{figure}

\paragraph{Expressive prompt corpus}
For our experiments that involve expressive prompt finetuning we generate a training corpus from UltraEdit dataset where each sample contains upto 5 expressive edits as shown in \ref{fig:Expressive generation}. Once these edits are generated offline, during training we sample 100k training data with expressive prompts and 100k samples without expressive prompts. In addition we add 200k samples of T2I to align with OmniGen style of training. Some examples of Expressive instructions that are more step-by-step edits to perform the core task is shown in \ref{fig:synthetic_expressive_prompts}. For our training we prompt Gemini Flash 2.0. Further during training we append each edit prompt in order shown in \ref{fig:Expressive generation}. OmniGen supports causal attention for text and we add expressive prompting in text space by appending each expressive prompt followed by a single pause token. The image id is added in the beginning along with the core instruction prompt.  

\begin{figure}[htbp]
\centering
\begin{tcolorbox}[
  title=Examples of Synthetic Expressive Prompts,
  colback=gray!3,
  colframe=gray!50!,
  fonttitle=\bfseries,
  width=\textwidth,
  boxrule=0.4pt,
  arc=3pt,
  outer arc=3pt,
  left=4pt,
  right=4pt,
  top=3pt,
  bottom=3pt,
  boxsep=3pt,
  coltitle=black,
  enhanced,
  sharp corners=south,
]
\scriptsize
\textbf{Instruction 1:} In the image \texttt{<img><|image\_1|></img>} change the pizza into a giant chocolate chip cookie.\\[2pt]
\textbf{Expressive Prompts:}
\begin{itemize}
  \item Identify the pizza in the image and determine its size and location.
  \item Create a new layer for the giant chocolate chip cookie and position it in the same location as the pizza.
  \item Use a texture or pattern overlay to give the cookie a realistic appearance.
  \item Adjust the size and shape of the cookie to match the pizza, ensuring it is large enough to replace the pizza.
\end{itemize}

\vspace{5pt}
\textbf{Instruction 2:} In the image \texttt{<img><|image\_1|></img>} replace the potatoes with cupcakes.\\[2pt]
\textbf{Expressive Prompts:}
\begin{itemize}
  \item Remove the potatoes from the image.
  \item Add cupcakes to the image, replacing the potatoes.
  \item Adjust the lighting and composition of the image to make the cupcakes the main focus.
  \item Apply a filter or effect to the image to make the cupcakes look appetizing and appealing.
\end{itemize}
\end{tcolorbox}
\caption{Synthetic editing instructions decomposed into expressive multi-step prompts to guide MAgD training.}
\label{fig:synthetic_expressive_prompts}
\end{figure}

\subsection{Benchmarks}
\paragraph{Emu-Edit} is a popular benchmark known for precise and high-fidelity manipulations. It focuses on detailed instruction-following to achieve highly controllable and realistic image alterations. The model aims to address the limitations of prior systems in executing fine-grained edits based on natural language commands. We evaluate along all tasks i.e Local, Global, Background, Style , Text , Add and Remove Tasks. 
\paragraph{MagicBrush:}  A popular manually annotated dataset for Instruction based image editing. Unlike the other benchmarks addressed here, it also offers masks, However on our evaluations we do  not consider masked image conditioning. We evaluate our model on single turn edits.
\paragraph{Ideabench (Professional editing benchmark)}: Ideabench is a specialized benchmark designed to assess AI models on "professional-grade" image editing capabilities. It features complex, dense instructions spanning images to image tasks and image to image tasks. We evaluate on IdeaBench as we observe Zero-shot abilities of the model on these tasks especially package rendering, Brand merchandise edits, Image id's transfer. 
\paragraph{Complex-Edit}: Complex-Edit a recent work is a meticulously curated benchmark of 2.4k Image -Instruction pairs. Each instruction is designed to be heirarchical with increasing complexity. Where, a long instruction is decomposed into sub-instructions. Often multiple subinstructions are augmented to obtain higher complexities. We perform show evaluations on Real datasets only although complex edit offers another subset of synthetic datasets. 

More details of task breakdowns in \ref{tab:Evaluation setup} and \ref{tab:editing_capabilities}


\begin{table}[htbp]
\centering
\caption{\textbf{Evaluation Setup}: We evaluate across four different benchmarks. \textsc{IdeaBench}, \textsc{MagicBrush}, and \textsc{Complex-Edit} involve real images, while \textsc{Emu-Edit} is a synthetically generated benchmark. Inference scaling results are reported on a randomly selected subset of datasets with proportionally similar task distributions.}
\label{tab:Evaluation setup}
\renewcommand{\arraystretch}{1.2}
\begin{tabular}{p{4.2cm}cccc}
\toprule
\rowcolor{gray!20}
\textbf{Metric} & \textbf{Emu-Edit} & \textbf{IdeaBench} & \textbf{MagicBrush} & \textbf{Complex-Edit (Real)} \\
\midrule
\multicolumn{5}{l}{\textbf{Experiments without Inference Scaling}} \\
Total test set size & $\sim$3k & 190 & 528 & 200 \\
\# Unique tasks & 8 & 15 & 7 & N/A \\
Complexity levels & N/A & N/A & N/A & 1,2,3,4 \\
\midrule
\multicolumn{5}{l}{\textbf{Experiments with Inference Scaling}} \\
\# Subset samples & 500 & N/A & 100 & 100 \\
Pause token lengths & 8, 16, 32 & N/A & 8, 16, 32 & 8, 16, 32 \\
\bottomrule
\end{tabular}
\end{table}

\begin{table}[htbp]
\centering
\caption{Editing Capabilities or tasks}
\label{tab:editing_capabilities}
\begin{tabular}{@{}>{\raggedright\arraybackslash}p{4cm}|l|l|l@{}} 
\toprule
\rowcolor{headergray}\textcolor{white}{} & \textcolor{white}{\textbf{Emu-Edit}} & \textcolor{white}{\textbf{MagicBrush}} & \textcolor{white}{\textbf{IdeaBench}} \\
\midrule
\rowcolor{rowgray}Region based & $\checkmark$ & $\checkmark$ & $\checkmark$ \\
\rowcolor{rowgray}Text editing & $\checkmark$ & $\times$ & $\checkmark$\\
\rowcolor{rowgray}Free form & $\checkmark$  & $\times$  & $\checkmark$ \\
\rowcolor{rowgray}Style editing & $\checkmark$  & $\times$  & $\checkmark$ \\
\rowcolor{rowgray}Image enhancements & $\times$  & $\times$  & $\checkmark$ \\
\bottomrule
\end{tabular}
\end{table}

\section{Qualitative Evaluations}
\begin{figure}[H] 
    \centering 
    \includegraphics[width=\textwidth]{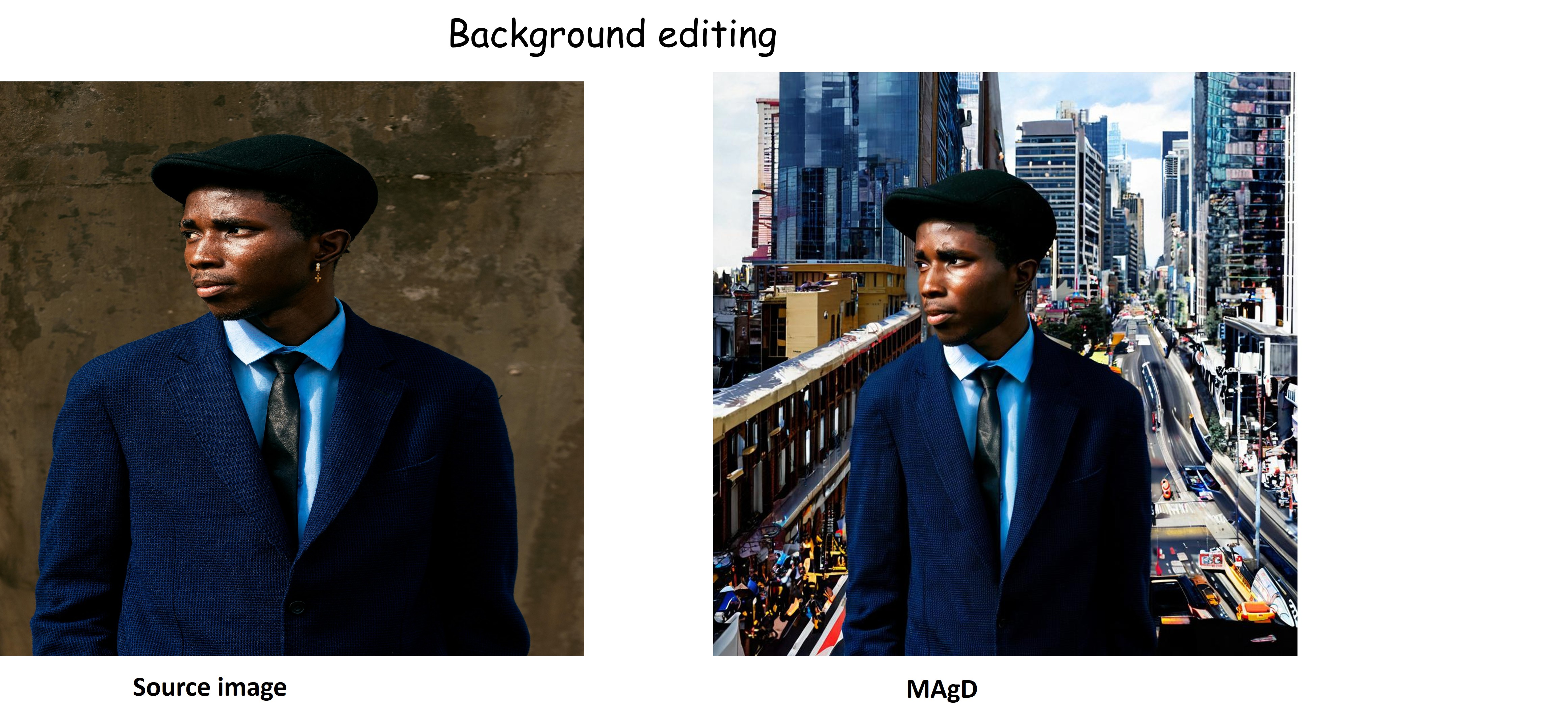} 
    \caption{\textbf{Prompt}: Change the background of this image featuring a man in a blue suit and a black hat. Replace the current plain, weathered wall with a modern cityscape, showing tall glass buildings and a busy street in the distance. Ensure the new background integrates smoothly with the lighting on the subject, keeping the city atmosphere vibrant and realistic..
}
    \label{fig:subfig1} 
\end{figure}
\begin{figure}[H]
        \centering
        \includegraphics[width=\textwidth]{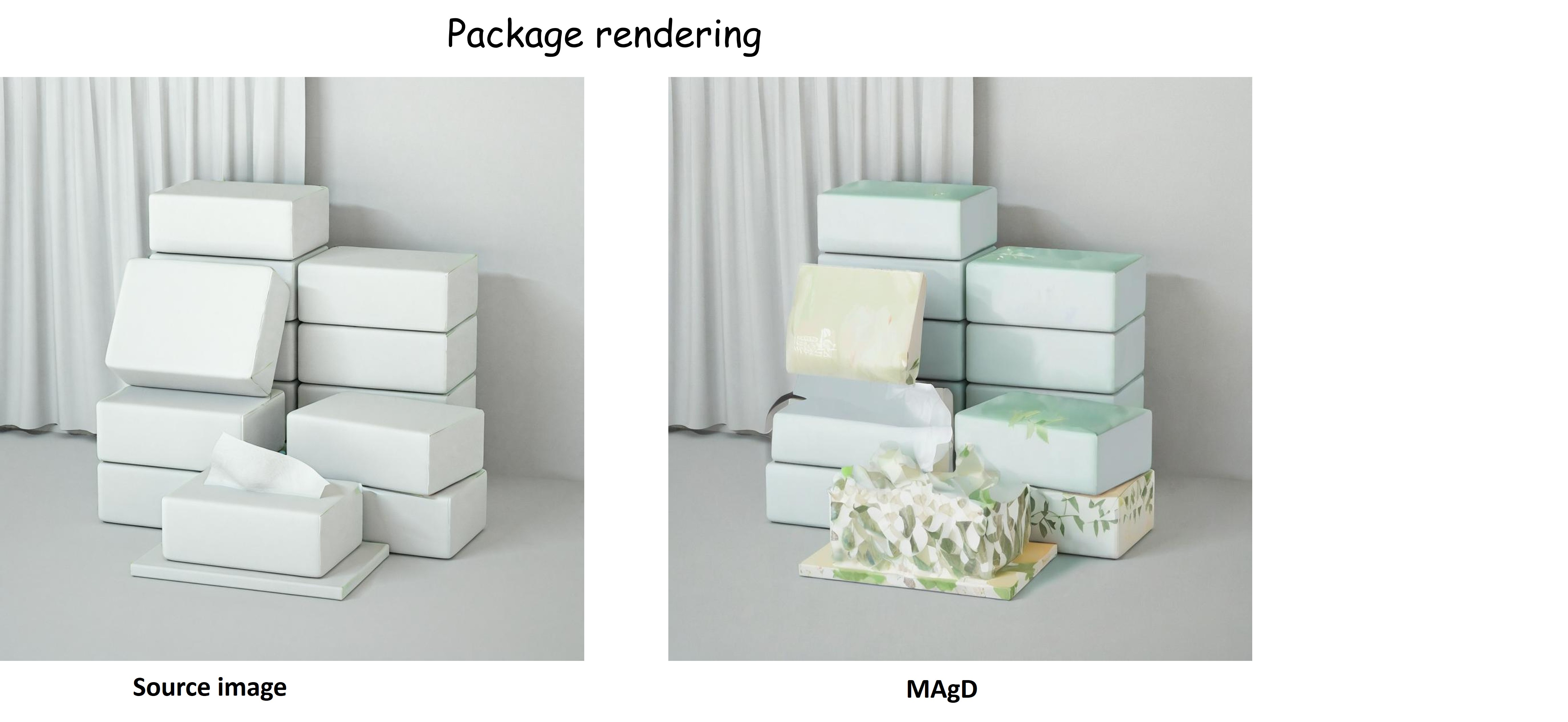} 
        \caption{\textbf{Ideabench Prompt}: Redesign the appearance of the square tissue boxes to reflect a clean and natural aesthetic. The body of the boxes should be a soft off-white color, adorned with light green leaf patterns, conveying a sense of nature and eco-friendliness. The surface of the boxes should be smooth, giving off a fresh and tidy texture. Each tissue box should retain a structured square shape with subtly rounded edges for a softer visual appeal. The top of each box should feature a tissue slot for easy extraction. The overall design should remain simple and harmonious, embodying a modern and fresh style with a light, natural touch.
}
        \label{fig:subfig2} 
\end{figure}

\begin{figure}[h]
        \centering
        \includegraphics[width=\linewidth]{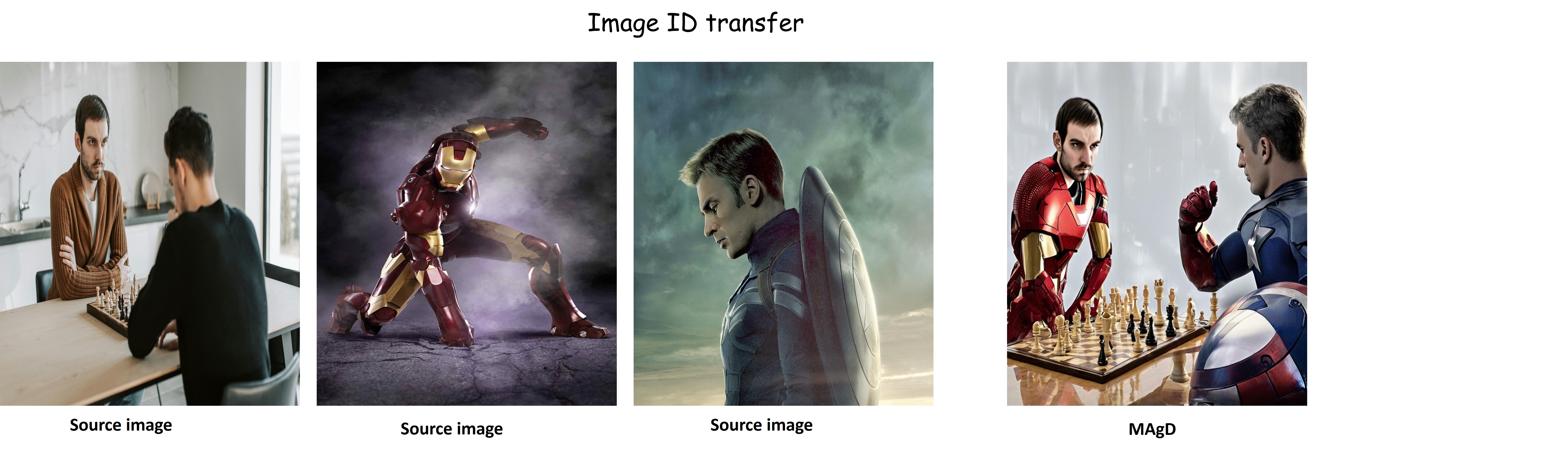} 
        \caption{\textbf{Ideabench Prompt for Images to image }: Generate an image where the two people playing chess in the first image are replaced by Iron Man from the second image and Captain America from the third image. Keep the primary elements of the original image, such as the chessboard, background, and furniture, unchanged. Ensure that Iron Man and Captain America's poses match the context of playing chess, adjusting their positions slightly if necessary. Their facial expressions should reflect concentration on the chess game.
}
        \label{fig:subfig2} 
\end{figure}



\begin{figure}[H] 
    \centering 

        \includegraphics[width=\textwidth]{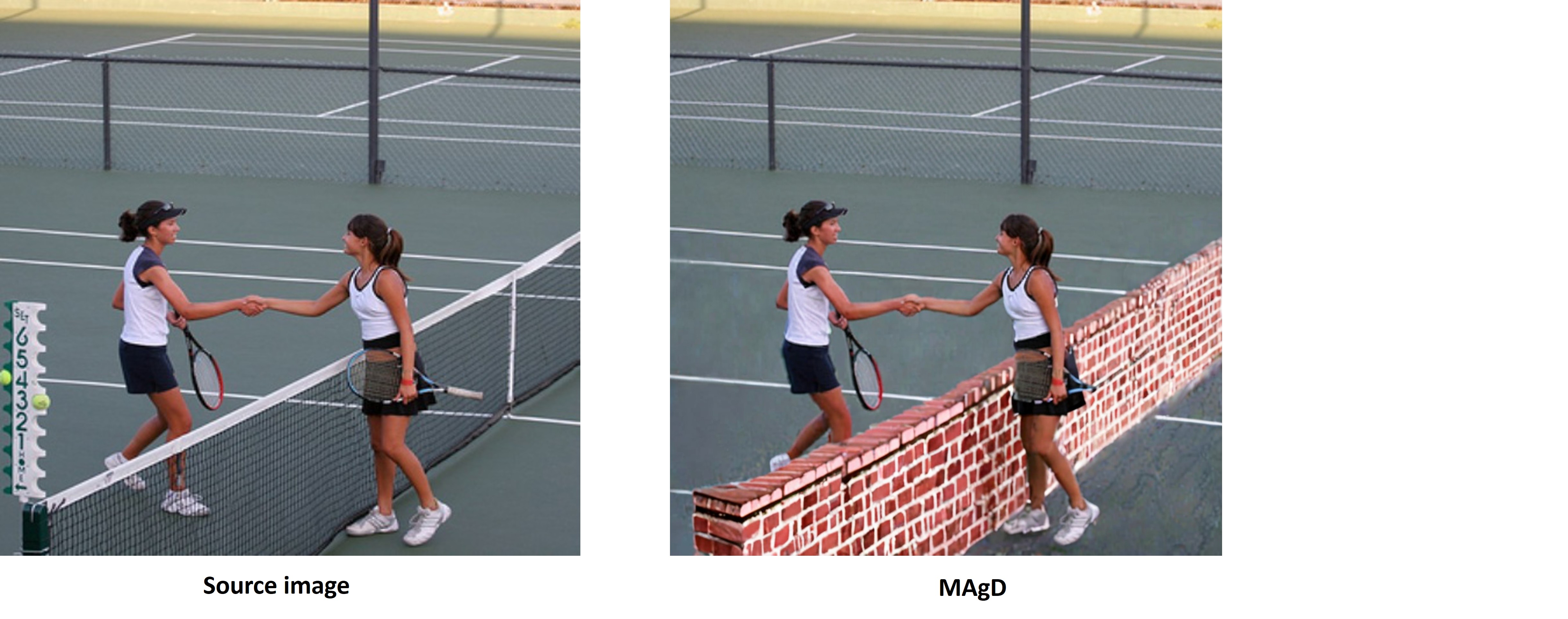} 
        \caption{Qualitative results on Emu-Edit for \textbf{Local Task}: "Replace the net with a brick wall"}
        \label{fig:subfig1} 
\end{figure}
\begin{figure}[H]
        \centering
        \includegraphics[width=\textwidth]{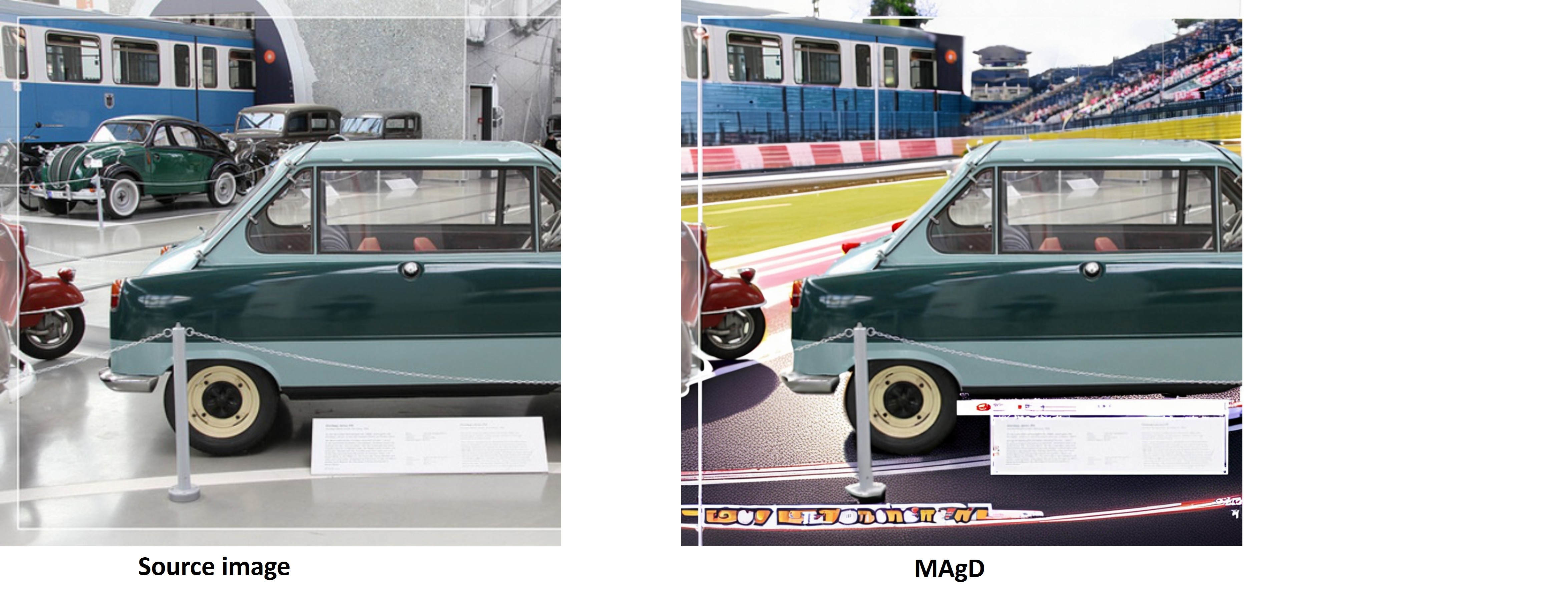} 
        \caption{Qualitative results on Emu-Edit for \textbf{Background Task}: "Make the background a race track"}
        \label{fig:subfig2} 
\end{figure}

\begin{figure}[H]
        \centering
        \includegraphics[width=0.85\textwidth]{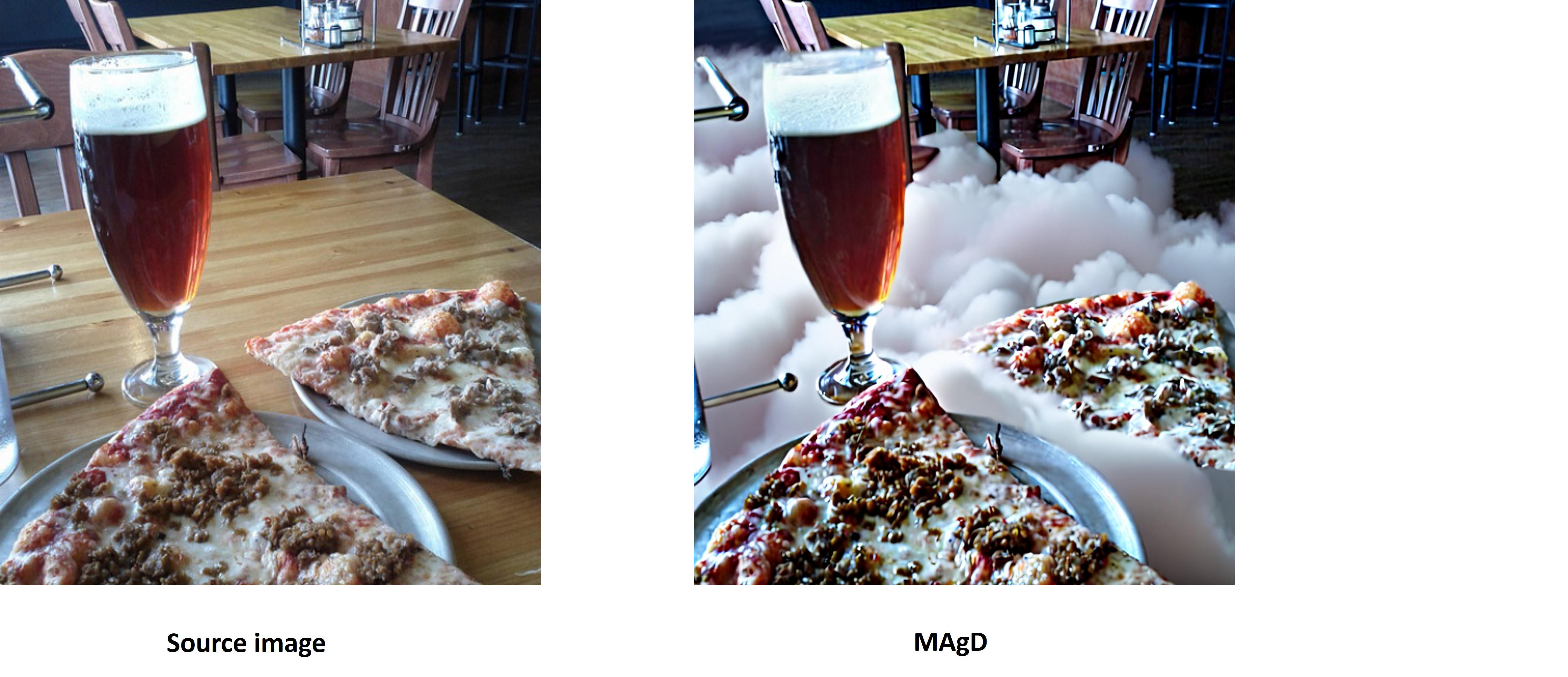} 
        \caption{Qualitative results on Emu-Edit for \textbf{Global Task}: "Set this to look like it is floating in the sky surrounded by white fluffy clouds"}
        \label{fig:subfig3} 
\end{figure}

\begin{figure}[htbp]
        \centering
        \includegraphics[width=0.85\textwidth]{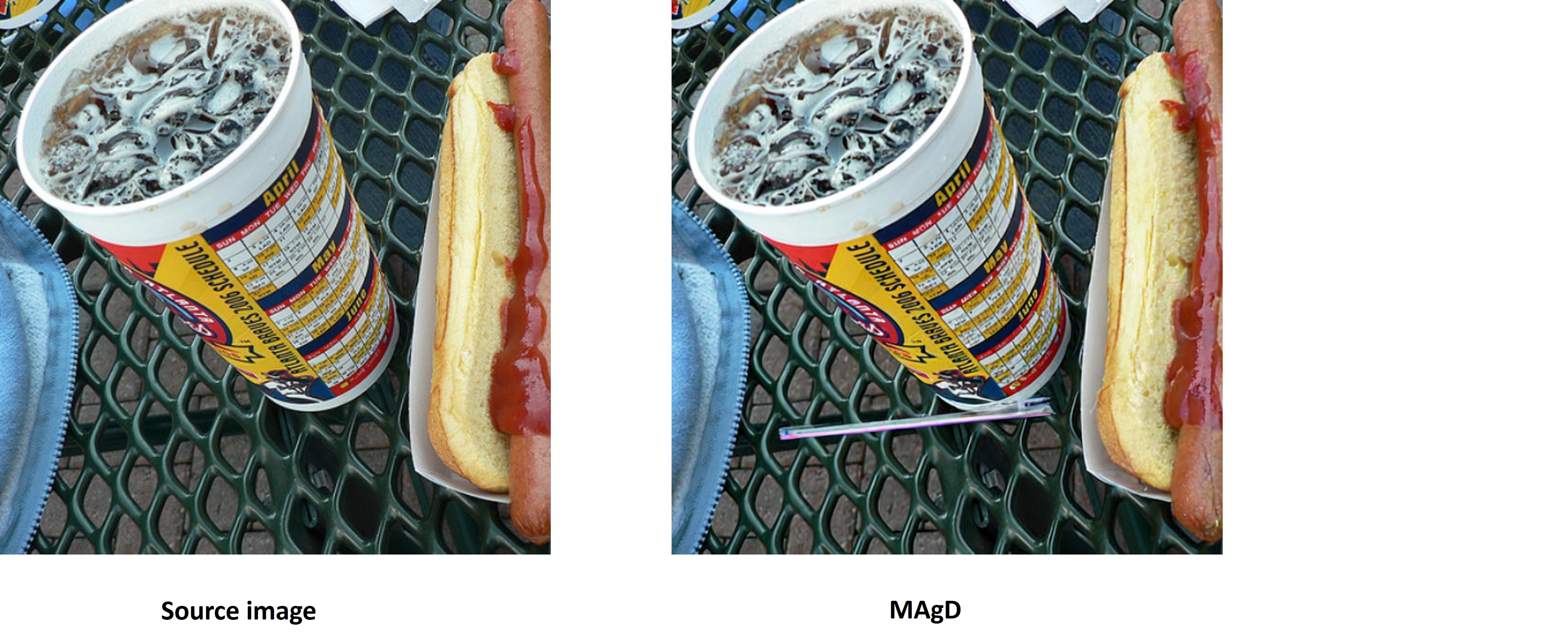} 
        \caption{Qualitative results on Emu-Edit for \textbf{Add task}: "Add a straw to the drink"}
        \label{fig:subfig4} 
\end{figure}

\begin{figure}[H]
    \centering
    \includegraphics[width=0.85\linewidth]{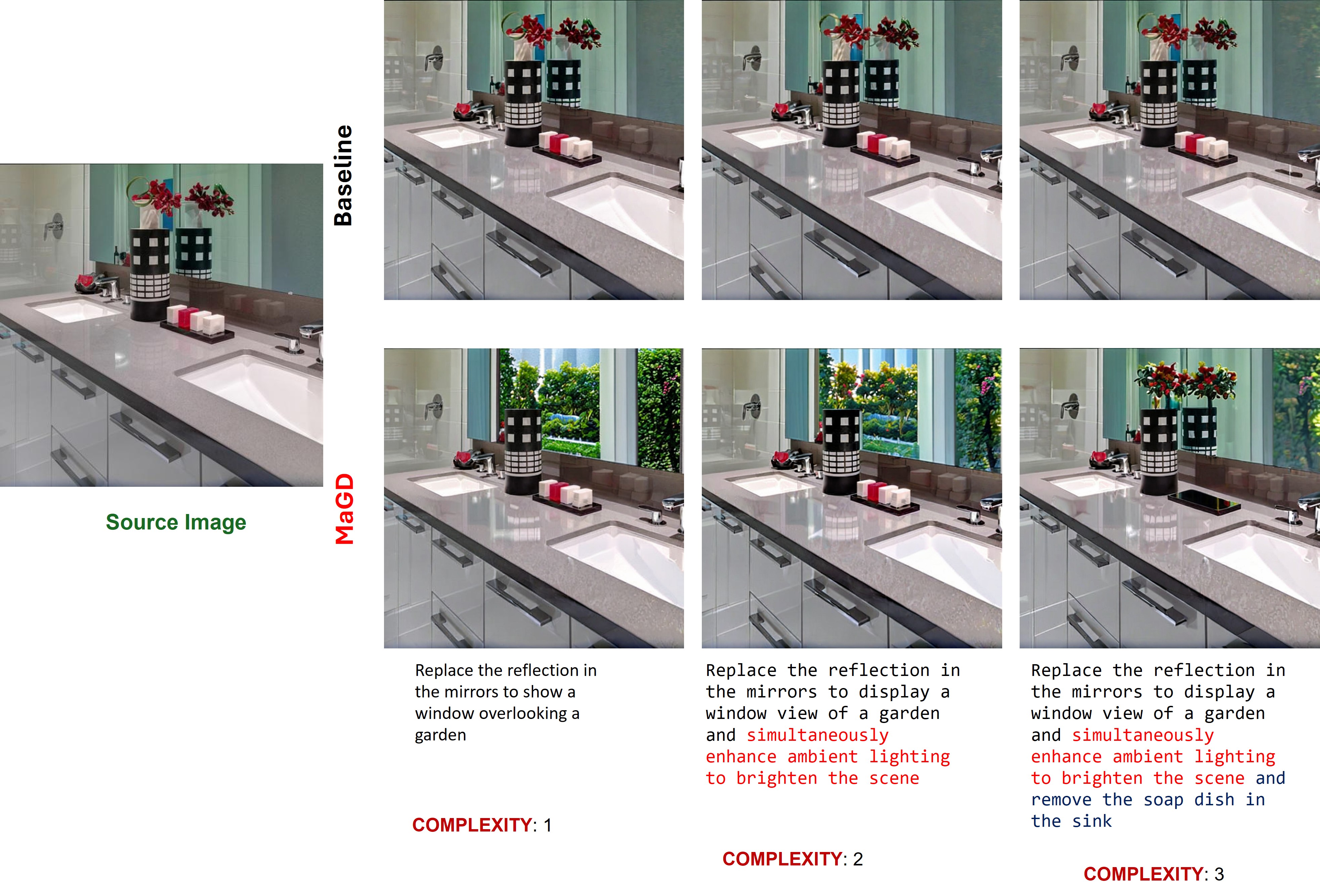}
    \caption{\textbf{Complex edit Evaluations} Comparisons between Finetuned and MAgD across different complexities}
    \label{fig:enter-label}
\end{figure}

\appendix



\newpage
\end{document}